%% file: main.tex
\documentclass[]{stem}
\usepackage{microtype}
\usepackage{amsfonts}
\usepackage{graphicx}
\usepackage{booktabs}
\usepackage{wrapfig}
\usepackage{amsmath}
\usepackage{amsthm}
\usepackage{subcaption}

\usepackage{algpseudocode}
\usepackage[linesnumbered,lined,boxed,commentsnumbered,ruled,longend]{algorithm2e}
\usepackage{mathtools}
\usepackage{arydshln}
\usepackage{seqsplit}
\usepackage{threeparttable}
\usepackage{tablefootnote}
\usepackage{makecell}
\usepackage[capitalize,noabbrev]{cleveref}
\theoremstyle{plain}

\theoremstyle{definition}

\theoremstyle{remark}

\usepackage{enumitem}
\newcommand{\wrapttt}[1]{\texttt{\seqsplit{#1}}}

\definecolor{lavenderpink}{rgb}{0.95, 0.85, 0.95}
\definecolor{softlavender}{rgb}{0.64, 0.50, 0.68}
\title{STEM: Scaling Transformers with Embedding Modules}

\usepackage{tikz}
\usetikzlibrary{arrows.meta,positioning,fit,calc,shapes.geometric}

\tikzset{
  box/.style={rounded corners=6pt, draw, thick, inner sep=4pt, fill=#1},
  op/.style={box=#1, minimum width=2.1cm, minimum height=0.8cm},
  tinyop/.style={box=#1, minimum width=1.8cm, minimum height=.65cm},
  io/.style={box=#1, minimum width=1.4cm, minimum height=.65cm},
  circ/.style={circle, draw, fill=white, thick, minimum size=6mm},
  note/.style={draw=gray, dashed, rounded corners=4pt, inner sep=3pt},
  flow/.style={-{Latex[length=1.2mm,width=0.9mm]}, semithick, shorten >=1pt},
  dashedflow/.style={densely dashed, semithick, -{Latex[length=1.2mm,width=0.9mm]}},
  module/.style={rounded corners=10pt, draw=black!65, blur shadow,
                 inner sep=8pt, fill=white},
  subop/.style={rounded corners=8pt, draw=black!55, thick, inner sep=6pt, fill=#1},
  dottedflow/.style={densely dotted, -Latex, semithick},
  pill/.style={rounded corners=10pt, draw=black!60, thick, fill=#1, inner sep=6pt},
  op/.style={rounded corners=8pt, draw=black!65, thick, fill=#1, inner sep=7pt, minimum height=12mm},
  circop/.style={circle, draw=black!70, thick, fill=white, minimum size=7mm},
  panel/.style={rounded corners=10pt, draw=#1!65!black, thick, fill=#1!5,
                align=left, inner sep=8pt, text width=0.34\textwidth},
  tag/.style={rounded corners=6pt, draw=black!40, thin, fill=black!5, inner sep=3pt, font=\scriptsize\sffamily},
}

\author[\dagger\ddagger]{Ranajoy Sadhukhan}
\author[\mathparagraph]{Sheng Cao}
\author[\dagger]{Harry Dong}
\author[\mathsection]{Changsheng Zhao}
\author[\mathsection]{Attiano Purpura-Pontoniere}
\author[\mathparagraph]{Yuandong Tian}
\author[\mathsection]{Zechun Liu\textsuperscript{*}}
\author[\dagger]{Beidi Chen\textsuperscript{*}}

\affiliation[\dagger]{Carnegie Mellon University}
\affiliation[\mathsection]{Meta AI}
\contribution[\ddagger]{Work done during internship at Meta AI}
\contribution[\mathparagraph]{Work done at Meta AI}
\contribution[*]{Co-supervisors}

\contact{\email{rsadhukh@andrew.cmu.edu}}
\contact{\email{rick.caos@gmail.com}}
\contact{\email{harryd@andrew.cmu.edu}}
\contact{\email{cszhao@meta.com}}
\contact{\email{attiano@meta.com}}
\contact{\email{yuandong.tian@gmail.com}}
\contact{\email{zechunliu@meta.com}}
\contact{\email{beidic@andrew.cmu.edu}}

\abstract{$\!\!$Fine-grained sparsity promises higher parametric capacity without proportional per-token compute, but often suffers from training instability, load balancing, and communication overhead. We introduce \textbf{STEM} (\emph{Scaling Transformers with Embedding Modules}), a static, token-indexed approach that replaces the FFN up-projection with a layer-local embedding lookup while keeping the gate and down-projection dense. This removes runtime routing, enables CPU offload with asynchronous prefetch, and decouples capacity from both per-token FLOPs and cross-device communication. Empirically, STEM trains stably despite extreme sparsity. It improves downstream performance over dense baselines while reducing per-token FLOPs and parameter accesses (eliminating roughly one-third of FFN parameters). STEM learns embedding spaces with large angular spread which enhances its \emph{\textbf{knowledge storage capacity}}. 
More interestingly, this enhanced knowledge capacity comes with \emph{\textbf{better interpretability}}. The token-indexed nature of STEM embeddings allows simple ways to perform \emph{knowledge editing} and \emph{knowledge injection} in an interpretable manner without any intervention in the input text or additional computation.
In addition, STEM strengthens long-context performance: as sequence length grows, more distinct parameters are activated, yielding practical test-time capacity scaling. Across 350M and 1B model scales, STEM delivers up to $\sim$\textbf{3--4\%} accuracy improvements overall, with notable gains on knowledge and reasoning-heavy benchmarks (ARC-Challenge, OpenBookQA, GSM8K, MMLU). Overall, STEM is an effective way of scaling parametric memory while 
providing \textbf{\textit{better interpretability}}, \textbf{\textit{better training stability}} and \textbf{\textit{improved efficiency}}.
}
\metadata[Github]{\url{https://github.com/Infini-AI-Lab/STEM}}
\metadata[Website]{\url{https://infini-ai-lab.github.io/STEM}}

\begin{document}

\maketitle
\input{sections/introduction}

\input{sections/method}
\input{sections/experiments}

\input{sections/analysis}

\input{sections/related_works}
\input{sections/conclusion}
\clearpage
\newpage
\bibliographystyle{assets/plainnat}
\bibliography{paper}
\clearpage
\newpage
\beginappendix
\input{sections/appendix}
\end{document}

%% file: sections/introduction.tex
\section{Introduction}

\begin{figure*}[h]
    \centering
    \subfloat[Validation PPL scores]{
        \includegraphics[width=0.30\linewidth]{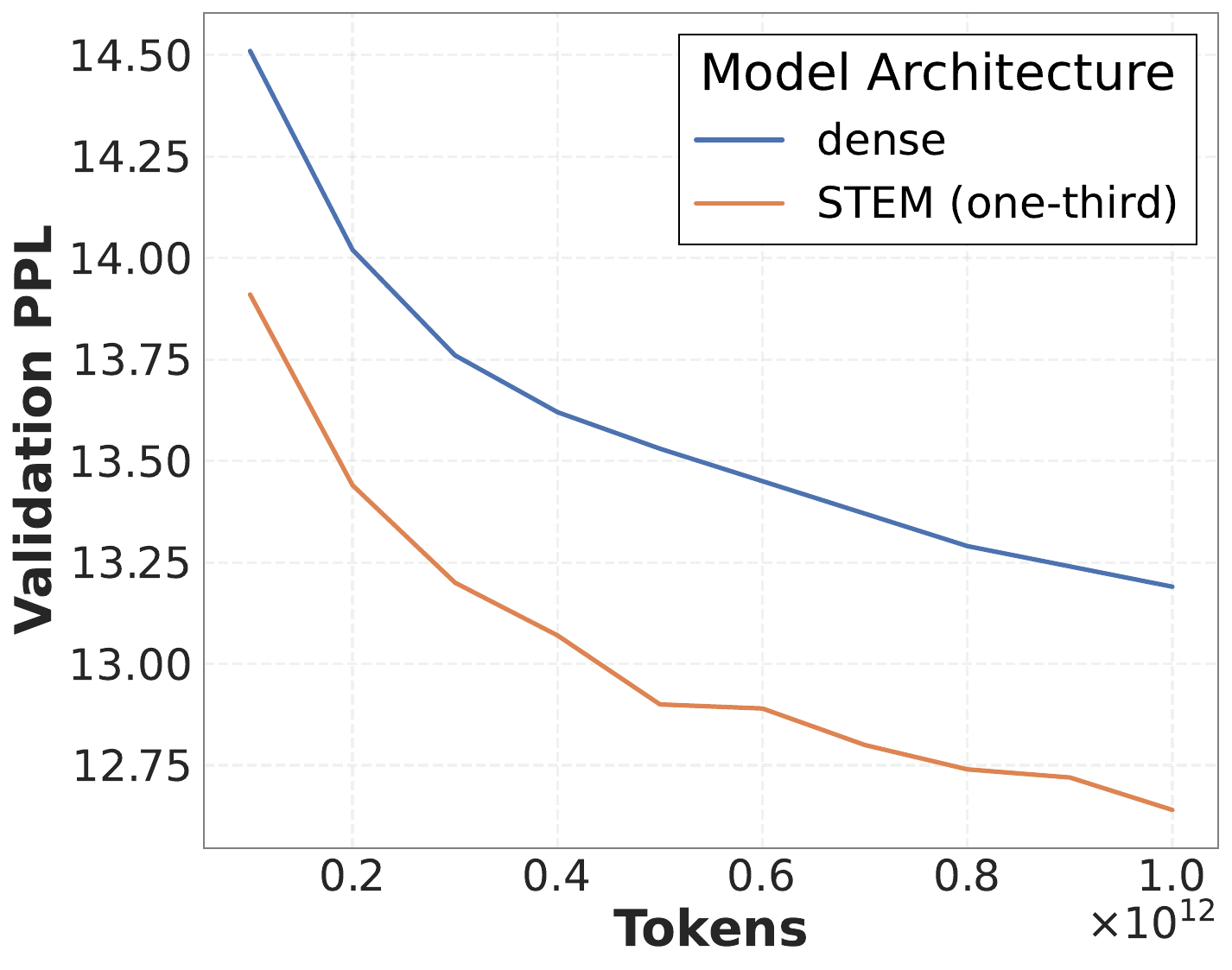}
        \label{fig:val_loss_1B}
    }
    \subfloat[Context Length Scalability]{
        \includegraphics[width=0.30\linewidth]{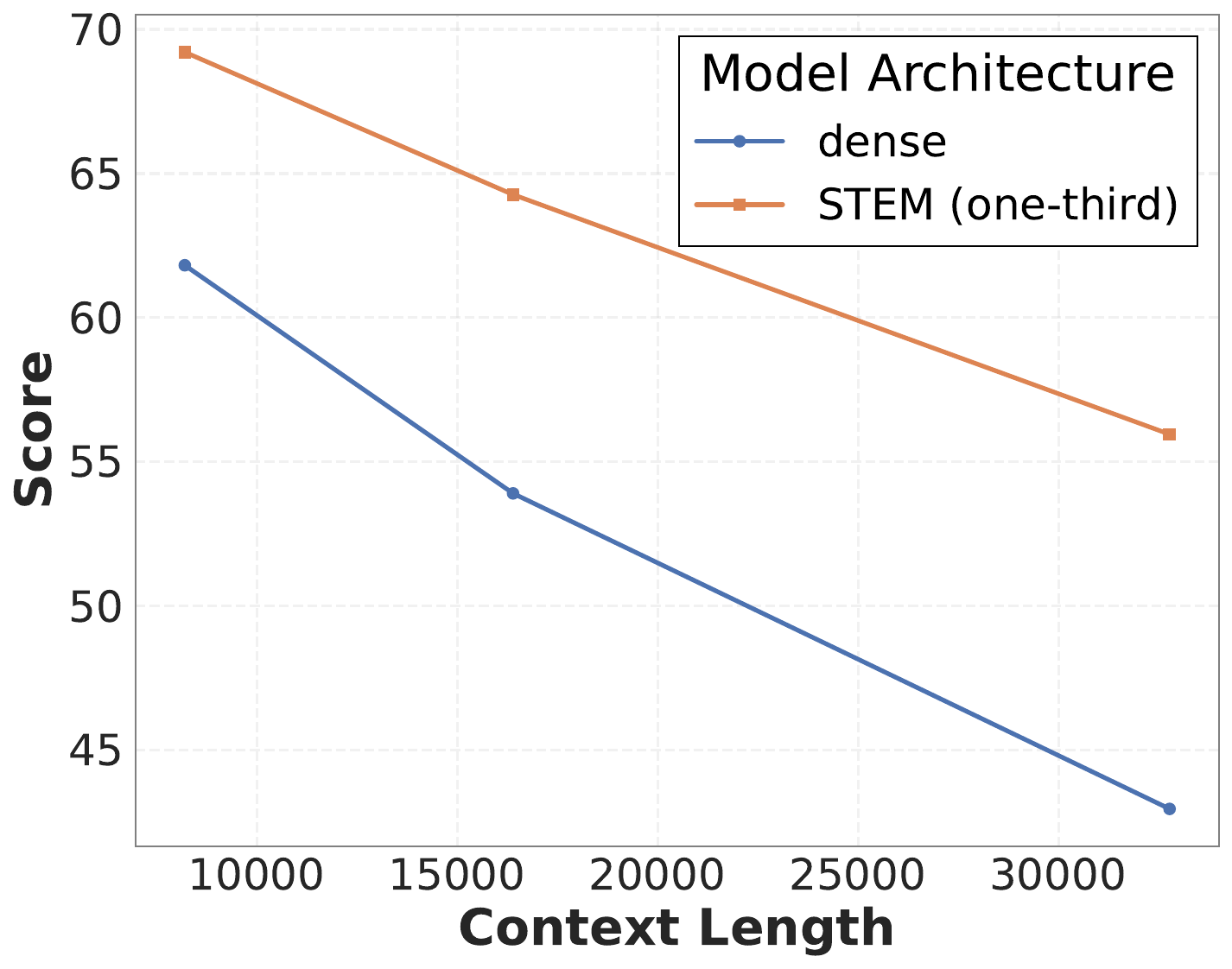}
        \label{fig:niah}
    }
    \subfloat[STEM layer]{
        \includegraphics[width=0.34\linewidth]{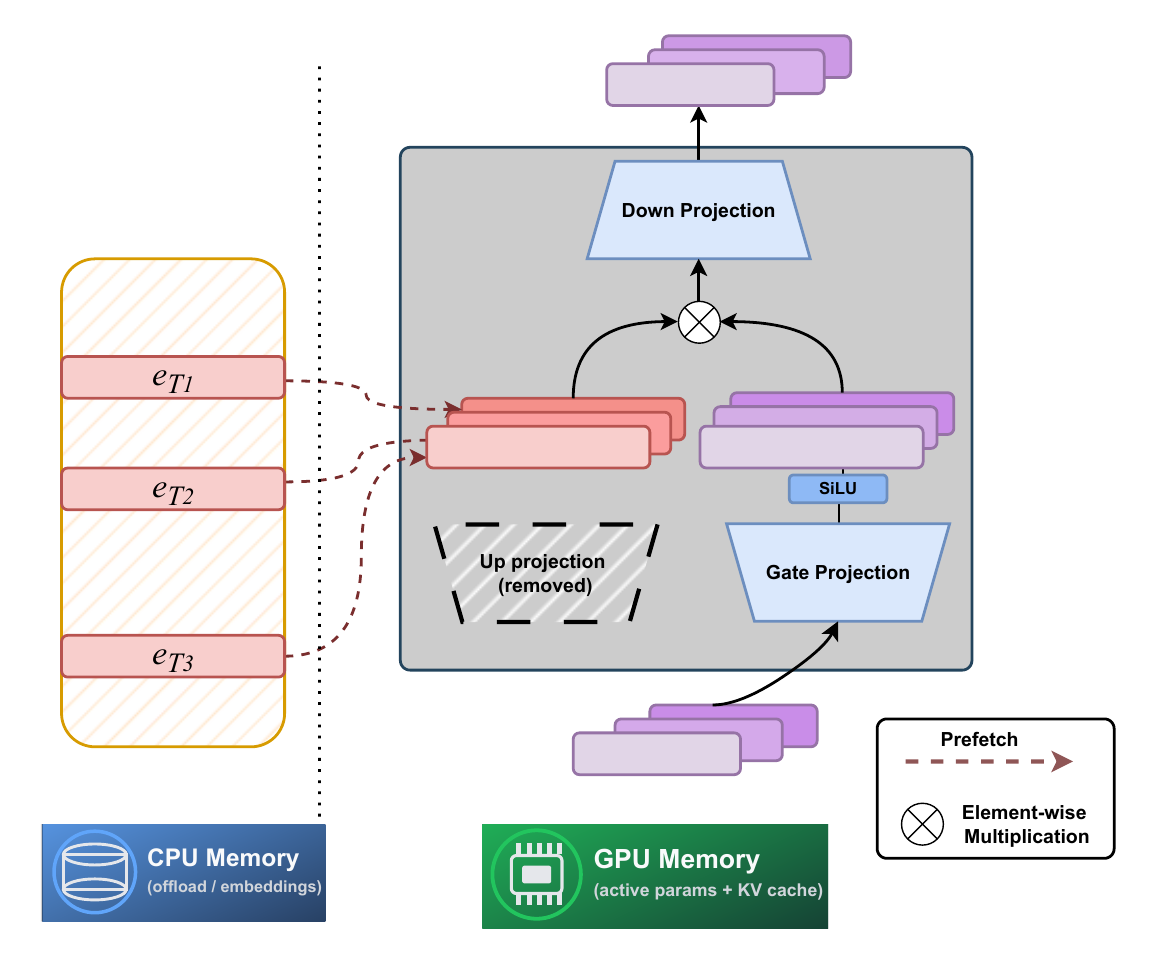}
        \label{fig:schematic}
    }
    \caption{(a) Validation PPL vs. training tokens for 1B STEM vs. dense; (b) Needle-in-a-Haystack at 8k/16k/32k; (c) STEM layer: embedding tables offloaded to CPU and token-indexed ones are prefetched to GPU.}
\end{figure*}

Sparse computation is a key mechanism for realizing the benefits predicted by parameter-scaling laws \citep{kaplan2020scalinglawsneurallanguage, hoffmann2022trainingcomputeoptimallargelanguage} without proportionally increasing per-token compute. In particular, Mixture-of-Experts (MoE) \citep{shazeer2017outrageouslylargeneuralnetworks, artetxe2022efficientlargescalelanguage, fedus2022switchtransformersscalingtrillion} models have been adopted in several frontier LLMs \citep{qwen3technicalreport, qwen3max, dai2024deepseekmoeultimateexpertspecialization} because they raise \emph{parametric capacity} at roughly constant \emph{activated} FLOPs by sparsely activating a small subset of experts per token. Recent work \citep{boixadsera2025powerfinegrainedexpertsgranularity, he2024mixturemillionexperts, dbrx2024, dai2024deepseekmoeultimateexpertspecialization} further advocate for \emph{finer-grained} sparsity that employs large number of \emph{micro-experts} to achieve better expressivity, enhanced knowledge storing capacity, and favorable efficiency metrics.

However, finer granularity introduces nontrivial challenges in both optimization and systems. On the training side, even large fraction of experts can remain under-trained \citep{huang2025ultrasparsememorynetwork} due to a highly non-uniform routing and result in training instability. While load-balancing objectives \citep{shazeer2017outrageouslylargeneuralnetworks, fedus2022switchtransformersscalingtrillion, lepikhin2020gshardscalinggiantmodels} can address these issues, they may interfere with the primary objective if not carefully tuned \citep{dai2024deepseekmoeultimateexpertspecialization, qiu2025demonsdetailimplementingload, go2025moetuneroptimizedmixtureexpert}. On the systems side, increasing the number of experts typically raises the number of all-to-all messages while shrinking message sizes, degrading bandwidth utilization and amplifying communication overhead \citep{huang2024toward, li2025speculativemoecommunicationefficient}. Finer granularity can also reduce parameter-access locality and degrade kernel efficiency when expert subnetworks become too small for dense linear-algebra kernels to reach high occupancy, yielding suboptimal end-to-end performance. Finally, these large-scale fine-grained sparse networks are far from interpretable. It is difficult to understand the roles of each micro-expert.
To harness the full potential of fine-grained sparsity, we require: \textbf{(a)} \emph{stable optimization}, \textbf{(b)} \emph{broad expert utilization} (each micro-expert learns useful representations), and \textbf{(c)} \emph{negligible expert-retrieval latency and communication overhead}. Additionally, we would like our sparse architecture to be \textbf{(e)} \emph{more interpretable} and ensure that we are using all the micro-experts in a more transparent manner.

We identify static sparsity as a potential solution to achieve these desired properties. 
Static sparsity keeps the compute path predictable (no runtime routing latency), enables prefetch and CPU offloading (removing the need for inter-node communication). 
Recently, static sparsity via token-indexed routing has emerged as a promising direction \citep{roller2021hashlayerslargesparse, gemma3n2024} with strong performance guarantees. Additionally, the token-indexed nature makes this static sparsity more interpretable as each micro-expert can correspond to a given token ID. 
However, such token-based selection strategy lacks context adaptivity. If applied naively, it can reduce the expressivity of the model and degrade quality despite more parameters. Our ablation study in sec. \ref{sec:stem placement} highlights the criticality of selecting the suitable module for sparsification.



Based on these observations, we introduce \emph{STEM}, a static, token-indexed, fine-grained mechanism that replaces \emph{only} the up-projection in gated FFNs with a token-specific vector retrieved from a layer-local embedding table. The gating and down-projection paths are preserved and shared across tokens. We observe that STEM achieves,


\underline{\textit{Better Training Stability:}} Despite being extremely sparse, STEM does not exhibit any training instability issues as usually seen in MoE models. Figure \ref{fig:comp_w_moe} shows that unlike MoE models, STEM does not exhibit any loss spikes. 

\underline{\textit{Improved Performance with Larger Knowledge Capacity:}} STEM learns a representation space for the embeddings that is conducive to better information storage. The learned embeddings exhibit a large angular spread (i.e., low pairwise cosine similarity), which reduces representational interference and improves addressability of the parametric memory. As a result, it effectively increases the distinct “slots” available for storing and retrieving information. In our downstream evaluation benchmark, STEM consistently outperforms the dense baseline on knowledge-intensive tasks like, ARC-Challenge \citep{allenai:arc}, and OpenBookQA \citep{OpenBookQA2018} by large margins ($\sim$9--10\%) and the improvement usually increases with more STEM layer inclusion. 

\underline{\emph{Interpretability features:}} Because each STEM embedding in every layer is tied to a specific token ID, individual “micro-experts’’ have clear, token-level semantics. This structure not only makes their role more interpretable, but also gives STEM models a surprising degree of controllability: by simply swapping the STEM table index (illustrated in Fig. \ref{fig:knowledge-combined}) while leaving the input text unchanged, we can systematically steer the model’s output distribution. Such interventions highlight how much factual knowledge is localized in these embeddings and how modular, editable, and attributable that knowledge becomes.

\underline{\textit{Improved Long-context Inference:}} During long-context inference, STEM activates more distinct parameters as sequence length grows, yielding test-time capacity scaling. As shown in Figure \ref{fig:niah}, the benefits strengthen with context: on Needle-in-a-Haystack (NIAH) \citep{kamradt_needle_2024}, the gap over the dense baseline increases from 8.4\% to 13\%. 

\underline{\textit{Training and Inference-time efficiency:}} STEM reduces both FLOPs as well as parameter loading cost by eliminating one-third of the parameters in FFN layers. Consequently, it is strictly more efficient during both computation-intensive training and prefilling, as well as in memory-intensive decoding. 



We benchmark STEM against the dense baseline with 350M MobileLLM \citep{liu2024mobilellmoptimizingsubbillionparameter} and Llama3.2-1B \citep{llama32_1b_meta_2024} model variants. Additionally, we compare with Hash Layer MoEs with the \emph{same total parameter count}. We report results on standard downstream suites across pretraining, mid-training, and context-length extension. STEM improves downstream accuracy by up to $\sim$3--4\% while reducing per-token FLOPs and parameter accesses by up to one-third. It also strengthens knowledge retrieval and mathematical reasoning, with gains on GSM8K \citep{cobbe2021gsm8k} and MMLU \citep{hendryckstest2021}, and shows pronounced improvements on Needle-in-a-Haystack \citep{kamradt_needle_2024} at longer contexts. Additionally, we illustrate the interesting ability of STEM to perform \emph{knowledge editing} with minimal intervention in Sec. \ref{subsec:knowledge-editing}.

%% file: sections/method.tex
\section{Background}
\begin{figure*}[t]
\centering

\begin{minipage}[t]{0.31\textwidth}
\centering
\begin{tikzpicture}[x=1pt,y=1pt, scale=0.6, every node/.style={transform shape}]
  \node[io=gray!30] (x) {Input $\mathbf{x}$};
  \node[op=teal!30, below=14pt of x] (Wg) {SiLU($\mathbf{W}^{g}(.)$)};
  \node[circ, below=14pt of Wg] (mul) {$\otimes$};
  \node[op=blue!30, left=14pt of mul] (Wu) {$\mathbf{W}^{u}(.)$};
  \node[op=violet!30, below=16pt of mul] (Wd) {$\mathbf{W}^{d}(\cdot)$};
  \node[io=gray!30, below=12pt of Wd] (y) {Output $\mathbf{y}$};

  \draw[flow] (x) -| (Wu);
  \draw[flow] (x) -- (Wg);
  \draw[flow] (Wu.east) -- (mul.west);
  \draw[flow] (Wg) -- (mul);
  \draw[flow] (mul) -- (Wd);
  \draw[flow] (Wd) -- (y);

  \node[below=2pt of y] {\textbf{(a) SwiGLU FFN}};
\end{tikzpicture}
\end{minipage}
\hfill
\begin{minipage}[t]{0.31\textwidth}
\centering
\begin{tikzpicture}[x=1pt,y=1pt, scale=0.6, every node/.style={transform shape}]
  \node[io=gray!30] (x) {Input $\mathbf{x}$};
  \node[op=orange!30, below=12pt of x] (router) {Router $r(\mathbf{x})$ (top-$r$)};

  \node[op=blue!20, below=20pt of router] (E2) {Expert $f_2$};
  \node[op=blue!20, left=22pt of E2] (E1) {Expert $f_1$};
  \node[op=blue!20, right=22pt of E2] (Ek) {$\cdots\; f_K$};

  \node[circ, below=18pt of E2] (comb) {$\sum$};
  \node[io=gray!30, below=12pt of comb] (y) {Output $\mathbf{y}$};

  \node[draw,dashed,rounded corners,inner sep=4pt,fit=(E1)(Ek)] (expbox) {};
  \node[font=\scriptsize, anchor=north east] at (expbox.north east) {};

  \draw[flow] (x) -- (router);
  \draw[flow] (router) -- (E1);
  \draw[flow] (router) -- (E2);
  \draw[flow] (router) -- (Ek);
  \draw[flow] (E1) -- (comb);
  \draw[flow] (E2) -- (comb);
  \draw[flow] (Ek) -- (comb);
  \draw[flow] (comb) -- (y);

  \node[below=2pt of y] {\textbf{(b) MoE FFN}};
\end{tikzpicture}
\end{minipage}
\hfill
\begin{minipage}[t]{0.31\textwidth}
\centering
\begin{tikzpicture}[x=1pt,y=1pt, scale=0.60, every node/.style={transform shape}]
  \node[io=gray!30] (x) {Input $\mathbf{x} (T)$};
  \node[op=teal!30,  below=14pt of x]                              (gate) { SiLU($\mathbf{W}^{g}(.)$)};
  \node[circ,        below=14pt of gate]            (mul)  {$\otimes$};
  \node[op=violet!30,below=14pt of mul]           (down) { $\mathbf{W}^{d}(.)$};
  \node[io=gray!30,  below=14pt of down]          (out)  {Output $\mathbf{y}$};

  \node[box=gray!15, fit=(gate)(mul)(down),
        inner sep=6pt, fill opacity=0.20]        (gpu)  {};
  \node[font=\scriptsize, above=0pt of gpu.north] {};

  \node[box=yellow!30, minimum width=1.6cm, minimum height=1.8cm,
        left=18pt of gpu.west, anchor=east]      (cpubox) {};
  \node[font=\scriptsize, above=0pt of cpubox.north] {CPU Embedding Table};
  \node[io=yellow!10, minimum width=1.2cm]       (emb1) at (cpubox.center) {$u_T$};

  \node[io=red!15, left=10pt of mul.west, anchor=east,
        minimum width=1.2cm]                     (pref) {$u_T$};

  \draw[dashedflow, red!70] (emb1.east) .. controls +(+8pt,0) and +(-8pt,0) .. (pref.west);
  \draw[flow] (pref.east) -- (mul.west);
  \draw[flow] (x) -- (gate);
  \draw[flow] (gate) -- (mul);
  \draw[flow] (mul) -- (down);
  \draw[flow] (down) -- (out);

  \node[below=2pt of out] {\textbf{(c) STEM}};
\end{tikzpicture}
\end{minipage}

\caption{Schematics of (a) SwiGLU FFN, (b) MoE FFN, and (c) STEM with a single prefetched token embedding. In MoE FFN, the full FFN module is considered as one expert.}
\label{fig:ffn_moe_stem_compact}
\end{figure*}
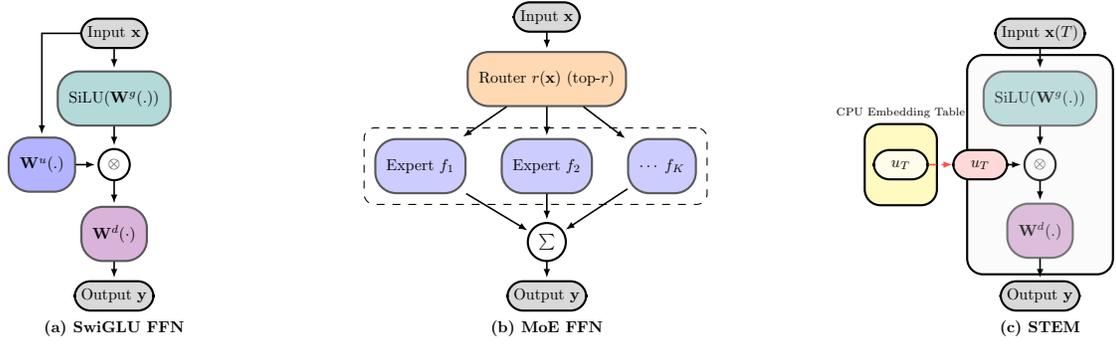

Consider a decoder-only transformer with $N$ layers, vocabulary size $V$, model width $d$, and feed-forward width $d_{\mathrm{ff}}$.
For a given layer $\ell$, the SwiGLU feed-forward block uses a gate projection $\mathbf{W}_{\ell}^{g}\!\in\!\mathbb{R}^{d_{\mathrm{ff}}\times d}$, an up projection $\mathbf{W}_{\ell}^{u}\!\in\!\mathbb{R}^{d_{\mathrm{ff}}\times d}$, and a down projection $\mathbf{W}_{\ell}^{d}\!\in\!\mathbb{R}^{d\times d_{\mathrm{ff}}}$. Consider, $t\!\in\!\{1,\dots,V\}$ denote the vocabulary id of the current token, and the corresponding input hidden state of the $\ell^{th}$ FFN layer is given by $\mathbf{x}_{\ell}\!\in\!\mathbb{R}^{d}$. Then the transformation in the FFN layer is 
\begin{equation}
  \mathbf{y}_{\ell}
  \;=\; \mathbf{W}_{\ell}^{d}\!\left(
      \mathrm{SiLU}\!\big(\mathbf{W}_{\ell}^{g} \mathbf{x}_{\ell}\big)
      \;\odot\;
      \big(\mathbf{W}_{\ell}^{u} \mathbf{x}_{\ell}\big)
  \right),
\end{equation}
where $\odot$ denotes elementwise multiplication.

\paragraph{Mixture-of-Experts (MoE).}
In MoE, a dense FFN is replaced by $K$ expert FFNs $\{f_{\ell,k}\}_{k=1}^K$ and a router $r_\ell(\mathbf{x}_\ell)$ that selects a small set $\mathcal{T}_\ell(\mathbf{x}_\ell)$ of top-$r$ experts with mixture weights $\pi_{\ell,k}(\mathbf{x}_\ell)$~\citep{artetxe2022efficientlargescalelanguage,fedus2022switchtransformersscalingtrillion}.
With SwiGLU experts,
\[
f_{\ell,k}(\mathbf{x}_\ell)
\coloneqq \mathbf{W}^{(d)}_{\ell,k}\!\left(\mathrm{SiLU}\!\big(\mathbf{W}^{g}_{\ell,k}\mathbf{x}_\ell\big)\;\odot\;
   \big(\mathbf{W}^{u}_{\ell,k}\mathbf{x}_\ell\big)\right),\quad
\mathbf{W}^{d}_{\ell,k}\in\mathbb{R}^{d\times d_{\mathrm{ff}}},
\]
the layer output is
\begin{equation}
  \mathbf{y}_{\ell}
  \;=\;
  \sum_{k \in \mathcal{T}_{\ell}(\mathbf{x}_\ell)}
  \pi_{\ell,k}(\mathbf{x}_\ell)\; f_{\ell,k}(\mathbf{x}_\ell).
\end{equation}

\paragraph{Hash-layer Mixture-of-Experts.}
To eliminate trainable routing and auxiliary losses, hash-layer MoE fixes a balanced, token-id–based mapping to experts~\citep{roller2021hashlayerslargesparse}. The FFN output becomes
\begin{equation}
  \mathbf{y}_{\ell}
  \;=\;
  \sum_{k \in \text{hash}(t)} f_{\ell,k}(\mathbf{x}_\ell).
\end{equation}

\paragraph{Scaling the number of experts.}
\emph{Mixture of Word Embeddings (MoWE)}~\citep{santos2023memoryaugmentedlanguagemodels} pushes expert granularity to the word level, instantiating tens to hundreds of thousands of small experts. This increases the model's \emph{knowledge capacity} and promotes \emph{word-specific specialization}. Similar to hash-layer MoE, MoWE uses a fixed mapping from token/word ids to expert subnetworks, enabling lightweight selection while retaining sparsity benefits. However, the extreme expert count magnifies two well-known MoE-related challenges:

\underline{\textit{Communication overhead.}}
Under expert parallelism, as the expert granularity increases, the peer-to-peer exchange becomes more fragmented, and communication becomes more \emph{latency-dominated} due to many small payloads and nontrivial packing/unpacking overhead. In practice, this raises end-to-end layer latency and reduces overlap with compute.

\underline{\textit{Unbalanced expert frequency.}}
Word frequencies are Zipfian, so a larger number of experts sharpens load skew: a few high-frequency experts receive disproportionate traffic while a long tail is rarely activated. This harms both statistical efficiency (slower or unstable learning for rare experts) and systems efficiency (capacity padding/drops, stragglers), further amplifying effective communication and synchronization costs during distributed training and inference~\citep{santos2023memoryaugmentedlanguagemodels}.


\paragraph{Per Layer Embedding.} Unlike MoWE, the Per Layer Embedding (PLE) \citep{gemma3n2024} share the gate projection and down projection of the FFN block across expert subnetworks. However, they do not completely dispense with the existing FFN block in each decoder layer. Instead, they complement the existing FFN with an additional PLE block. This provides token-level specificity at a small additional cost. 
Unlike MoWE, the PLE tables are not sharded across multiple devices. They are stored in node-local CPU memory instead and prefetched as required. Thus they avoid the high all-to-all communication traffic. In addition, by sharing the gate projection and down projection matrices, the negative effects of expert frequency mismatch are also ameliorated.

\section{Method}
\subsection{STEM}
\label{sec: defintion}
STEM builds upon the design of PLE and further explores the true potential of these layer-wise embedding tables in terms of both accuracy and efficiency. The STEM design can be expressed as follows,

For layer $\ell$, let $\mathbf{U}_{\ell}\!\in\!\mathbb{R}^{V\times d_{\mathrm{ff}}}$ be the per layer embedding table.
Given input $\mathbf{x}_{\ell}\!\in\!\mathbb{R}^{d}$, the STEM layer computes
\begin{equation}
  \mathbf{y}_{\ell}
  \;=\; \mathbf{W}_{\ell}^{(d)}\!\left(
      \mathrm{SiLU}\!\big(\mathbf{W}_{\ell}^{(g)} \mathbf{x}_{\ell}\big)
      \;\odot\;
      \mathbf{U}_{\ell}[t]
  \right),
\end{equation}
where $\mathbf{U}_{\ell}[t]\!\in\!\mathbb{R}^{d_{\mathrm{ff}}}$ is the row of $\mathbf{U}_{\ell}$ corresponding to token $t$ and $\odot$ denotes elementwise multiplication. 

It is important to note that, STEM makes some important design choices different from PLE. 

\begin{enumerate}
    \item PLE does not completely dispense with the existing FFN block in each decoder layer. Instead, the PLE FFN block is used as an \emph{additional component} with the regular FFN block in each PLE decoder layer. Thus, they increase additional overhead and effective layer depth of the model.

    \item PLE embedding tables are usually much more low-dimensional compared the intermediate dimension of the regular FFN layers. For instance, \texttt{gemma-3n-E4B-it}\citep{gemma3n2024} uses an FFN intermediate dimension of 16384, but a PLE dimension of just 256.
\end{enumerate}

To verify the sufficiency of the STEM embedding table, we also conduct ablation study with additional architectures which we discuss in detail in \ref{sec: additional arch}.

\subsection{Insights}
The following insights encourage us to make an attempt to realize the full potential of the layer-specific embedding tables unlike PLE.

\paragraph{Key–value memory view of FFNs.}
\label{sec: memory view}
To follow the motivation behind STEM, it is important to view FFN from a key-value memory perspective\citep{geva2021transformerfeedforwardlayerskeyvalue, meng2022locating}. A two-projection FFN can be read as a content-addressable key–value (KV) memory: with an input $\mathbf{x}\!\in\!\mathbb{R}^d$, hidden width $d_{\mathrm{ff}}$, and nonlinearity $\phi$, the block $\mathbf{y}= \mathbf{W}^{d}\phi(\mathbf{W}^{u}\mathbf{x})$ retrieves
\[
\mathbf{y} \;=\; \sum_{i=1}^{d_{\mathrm{ff}}} \underbrace{\phi(\langle \mathbf{k}_i,\mathbf{x}\rangle)}_{\text{addressing weight } \alpha_i(\mathbf{x})}\, \underbrace{\mathbf{v}_i}_{\text{value}},
\quad
\mathbf{k}_i \text{ is the $i$th row of } \mathbf{W}^{u},\;
\mathbf{v}_i \text{ is the $i$th column of } \mathbf{W}^{d}.
\]
Here, rows of $\mathbf{W}^{u}$ act as \emph{keys} scoring $\mathbf{x}$, while columns of $\mathbf{W}^{d}$ are \emph{values}; $\phi$ shapes the sparsity/selectivity of the addressing (e.g., ReLU for hard gating, GELU for soft gating). Gated linear units (GLUs) enrich this memory by factorizing the addressing into \emph{content} and \emph{gate} streams:
\[
\mathbf{y} \;=\; \mathbf{W}^{(d)}\!\big(\,(\mathbf{W}^{(u)}\mathbf{x}) \odot \sigma(\mathbf{W}^{g}\mathbf{x})\,\big)
\;=\; \sum_{i=1}^{d_{\mathrm{ff}}} \big\{ \underbrace{\langle \mathbf{k}_i,\mathbf{x}\rangle}_{\text{content}}\!\cdot\!\underbrace{\sigma(\langle \tilde{\mathbf{k}}_i,\mathbf{x}\rangle)}_{\text{gate}}\big\}\,\mathbf{v}_i,
\]
where $\mathbf{W}^{g}$ provides a second set of keys $\tilde{\mathbf{k}}_i$ that modulate each memory slot’s participation. This multiplicative interaction implements \emph{query-dependent, per-slot amplification/suppression}, yielding sharper, context-adaptive retrieval than a single-stream FFN. SwiGLU replaces the gate nonlinearity with $\mathrm{SiLU}$.

\paragraph{STEM design choice.} The memory view of FFN has been important for us to derive motivation for our final STEM design. To empirically verify the efficacy of the design described in Section \ref{sec: defintion}, we attempt to replace both up projection and gate projection independently using STEM embedding table. The down projection can not be replaced by an embedding module as it would break the model's forward path. As demonstrated in Table \ref{tab:pretrain-eval}, replacing gate projection hurts the downstream performance while replacing up projection enhances it. This can be explained by the roles that each of these matrices play in the FFN block according the memory view. While up projection generates the address for feature lookup in the down projection, gate projection provides context-dependent modulation to it for more effective information retrieval. Replacing gate projection with a context-agnostic embedding can impair its ability instead. For this reason, STEM replaces the up projection with the layer-wise embedding table.

This design choice allows STEM to enjoy benefits in terms of both performance and efficiency. Additionally, STEM offers an additional advantage in the form of better interpretability through better knowledge attribution. We explain the insights behind these benefits and empirically verify their efficacies.

\subsubsection{Better Information Storage Capacity}
Under the key–value memory view of FFNs (Section~\ref{sec: memory view}), the up-projection matrix maps each input hidden state to an \emph{address vector} that retrieves the relevant information from the subsequent down-projection. Although these address vectors live in a high-dimensional space, their intrinsic dimensionality is often much lower. In practice, FFN layers rely on mechanisms such as superposition to encode a large number of concepts within a relatively low-dimensional address space.

In contrast, STEM does not depend on an up-projection matrix to generate address vectors. Instead, the STEM embeddings themselves serve as token-specific address vectors, which are then modulated by the context-dependent gate projection output. The goal is to learn token-specific address vectors with as little mutual coherence as possible. Empirically, we observe that after training, the STEM embedding space exhibits a significantly larger angular spread than the address vectors produced by a standard FFN, as illustrated in Fig \ref{fig:angular spread}. We hypothesize that this reduced redundancy in the address space enables more precise and disentangled knowledge attribution, thereby improving the model’s effective information storage capacity.

\begin{figure*}[t]
  \centering
    \includegraphics[width=\linewidth,keepaspectratio]{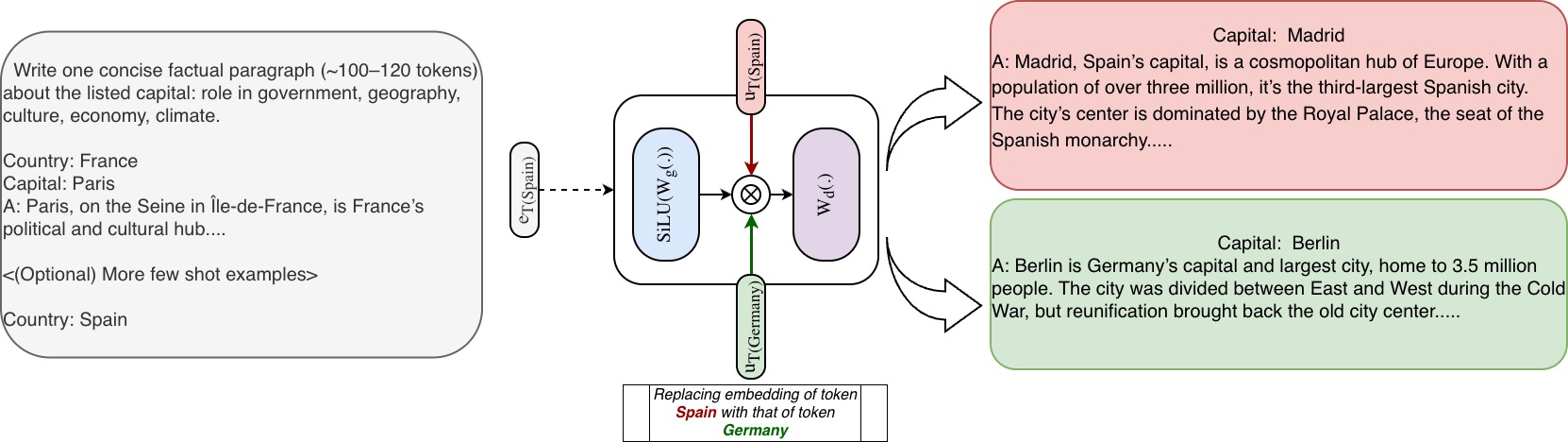}

  \caption{\textbf{Knowledge injection/edit demonstration.} Input text remains the same (\emph{Country: Spain}), but internally the PLE used for the token is swapped from Spain to Germany, flipping the generated capital/paragraph from Madrid to Berlin.}
  \label{fig:knowledge-combined}
\end{figure*}

\subsubsection{Knowledge Specificity \& Interpretability}
The STEM embeddings are attributed to the individual tokens. According to the memory view, this embedding should localize the necessary information associated with the corresponding token. 
Thus it can potentially function as steering vectors that can steer the output probability distribution based on careful modifications in the STEM embeddings in each layer without any modification in the text input. For instance, Figure \ref{fig:knowledge edit} illustrates how a minimal surgical modification in each STEM embedding layer can meaningfully modify the generation input, given the same input token.

 This direct knowledge attribution characteristics is missing in the standard FFN layers. Although recent works like sparse autoencoders \citep{huben2024sparse}, causal intervention \citep{meng2022locating} have tried to interpret the FFN modules, they usually require additional computation which can be prohibitively large. However, this knowledge attribution feature is inherently present in STEM and thus it attempts to address the longstanding tradeoff in Machine Learning between model \textit{performance} and \textit{interpretability}.

\subsubsection{Efficiency}
\label{sec:theory-efficiency}
\begin{table}[t]
\centering
\small
\caption{Theoretical efficiency for each decoder FFN layer when replacing the FFN up-projection with a token-indexed STEM embedding table. We assume SwiGLU, ignore biases, and count elementwise ops as $\mathcal{O}(DL)$.}
\label{tab:ple_theory}
\setlength{\tabcolsep}{6pt}
\begin{tabular}{lccc}
\toprule
& \textbf{FFN} & \textbf{STEM} & Savings ($\Delta$) \\
\midrule
\multicolumn{4}{l}{\emph{Prefill / training (batch size $B$, sequence length $L$)}} \\
\hdashline
\noalign{\vspace{0.2em}}
FLOPs & $B (3d_{\mathrm{ff}}dL + d_{\mathrm{ff}}L)$ & $B(2d_{\mathrm{ff}}dL + d_{\mathrm{ff}}L)$ & $B(dd_{\mathrm{ff}}L)$ \\
Communication & $0$ & $\mathrm{uniq}(BL) d_{ff}$ &  \\
\midrule
\multicolumn{4}{l}{\emph{Decoding (per step, batch size $B$)}} \\
\hdashline
\noalign{\vspace{0.2em}}
Parameter loading cost & $3dd_{\mathrm{ff}}$ & $2dd_{\mathrm{ff}}$ & $dd_{\mathrm{ff}}$ \\
Communication & $0$ & $B_{\text{uniq}}d_{ff}$ &  \\
\bottomrule
\label{tab:efficiency}
\end{tabular}\\
\vspace{3pt}
\raggedright\footnotesize
\textbf{Notation:} $d$: model width; $D$: FFN hidden size; $L$: context length; $L_{\text{uniq}}$: number of unique tokens in the $L$-token context; $B_{\text{uniq}}$: number of unique tokens across the batch at a decode step ($\le B$); $\mathrm{uniq}(BL)$: number of unique tokens across the $BL$ tokens in a training batch. \\
\textbf{Notes:} Training multiplies both FLOP counts by $\approx$ the usual forward+backward factor, but the saving $\Delta \text{FLOPs}=dDL$ remains. Communication doubles during training as gradients of the STEM embeddings are transferred back to CPU for optimizer update.
\end{table}
STEM improves both computation and memory access. During compute-intensive phases (training and prefill), replacing the FFN up-projection with token-indexed embeddings reduces the per-layer FLOPs. During memory-intensive decoding, it lowers parameter traffic relative to a dense up-projection. Table~\ref{tab:efficiency} summarizes the per-layer counts and the resulting savings. Below we present a simple theoretical analysis of the training and inference efficiency for a \emph{single} decoder layer.
\vspace{-0.25em}
\paragraph{Training efficiency.}
Consider a batch of $B$ sequences with sequence length $L$, hidden width $d$, and FFN hidden size $d_{\mathrm{ff}}$. Ignoring elementwise ops and biases, the per-layer training FLOPs (forward + backward) can be written as
\vspace{-0.25em}
\begin{align*}
F_{\text{train}}^{\text{base}}
&= B\bigl(4Ld^2 + 2L^2 d + 3 L d\, d_{\mathrm{ff}}\bigr), \\
F_{\text{train}}^{\text{stem}}
&= B\bigl(\underbrace{4Ld^2 + 2L^2 d}_{\text{Attn}} + \underbrace{2 L d\, d_{\mathrm{ff}}}_{\text{FFN}}\bigr).
\end{align*}
The per-layer FLOPs reduction of STEM is therefore
\[
\Delta F_{\text{train}}
= F_{\text{train}}^{\text{base}} - F_{\text{train}}^{\text{stem}}
= B L d\, d_{\mathrm{ff}},
\]
and the corresponding saving fraction is
\[
\text{saving fraction}
= \frac{\Delta F_{\text{train}}}{F_{\text{train}}^{\text{base}}}
= \frac{d_{\mathrm{ff}}}{4d + 2L + 3 d_{\mathrm{ff}}}.
\]
Plugging in the architecture hyperparameters for each Qwen2.5 model yields saving fractions of $21.7\%$ for Qwen2.5-1.5B, $22.8\%$ for Qwen2.5-3B, $23.9\%$ for Qwen2.5-7B, $19.7\%$ for Qwen2.5-14B, and $24.8\%$ for Qwen2.5-32B.

\paragraph{Inference efficiency.}
Prefill efficiency closely matches training efficiency because both are compute-bound. In contrast, decoding is primarily memory-bound: the dominant cost is loading parameters and KV cache rather than doing FLOPs. For a batch size $B$ and context length $L$, we can write the per-layer memory access cost as
\vspace{-0.25em}
\begin{align*}
M_{\text{dec}}^{\text{base}}
&= B\bigl(4d^2 + 2L d + 3 d\, d_{\mathrm{ff}}\bigr), \\
M_{\text{dec}}^{\text{stem}}
&= B\bigl(\underbrace{2L d}_{\text{KV cache}} + \underbrace{4d^2 + 2 d\, d_{\mathrm{ff}}}_{\text{projection params}}\bigr).
\end{align*}
\vspace{-0.25em}
The reduction in parameter loading cost is
\[
\Delta M_{\text{dec}}
= M_{\text{dec}}^{\text{base}} - M_{\text{dec}}^{\text{stem}}
= B d\, d_{\mathrm{ff}},
\]
so the saving fraction is
\[
\text{saving fraction}
= \frac{\Delta M_{\text{dec}}}{M_{\text{dec}}^{\text{base}}}
= \frac{d_{\mathrm{ff}}}{4d + 2L + 3 d_{\mathrm{ff}}},
\]
which matches the FLOPs saving factor during training and prefill. As the batch size grows, the linear layers become increasingly compute-bound, and STEM’s per-layer FLOPs reduction ensures that this efficiency gain is sustained even in the high-throughput regime.

A key difference from MoE is how cost scales with batch size. In STEM, parameter traffic grows mainly with the number of unique tokens seen. 
In contrast, MoE expert selection expands with batch size and routing diversity; larger batches tend to light up more experts, quickly eroding the sparsity benefit.

\subsubsection{VRAM and Communication Savings}
MoE models use a lot of VRAM. The expert subnetworks must stay on the GPU, or be fetched repeatedly. Expert parallelism also needs all-to-all communication, even when only a few experts are active~\citep{huang2024toward,go2025moetuneroptimizedmixtureexpert}.
STEM avoids these costs. Its embeddings are token-indexed and local to each layer, so the model can prefetch them without any routing logic. These tables are separate from the matmul weights, so we can offload them to CPU memory. In our setups, this frees up roughly one-third of the FFN parameter memory. We can also replicate the embedding tables in CPU memory on every serving node. This eliminates cross-node expert traffic and the synchronization overhead of expert parallelism. 
\vspace{-0.25em}
\paragraph{Prefetching cost.} The prefetching cost can be greatly reduced by deduplicating the STEM embeddings of the batched tokens. We can further cut traffic by caching the most frequently used STEM embeddings, using the extra memory we save from removing the up-projection matrices. As the model embedding size grows, compute cost increases quadratically, but prefetching cost grows only linearly. This makes CPU-offloaded STEM increasingly attractive and scalable for larger model sizes.

\subsubsection{Context-length Adaptive Parameter Usage}
\label{sec:ctx-adaptive}
Because \textsc{STEM} employs token-indexed, fine-grained sparsity, the number of \emph{distinct} parameters touched in a forward pass grows with the number of \emph{unique} tokens in the window. Aside from the shared projections in attention (Q/K/V/O) and the gated FFN’s gate/down projections, the STEM module draws one vector per token ID per layer; repeated tokens reuse the same vector, while novel tokens activate new ones. Let $L$ be the context length and $L_{uniq}$ the count of unique token ids in the sequence; with STEM applied at layers $\mathcal{S}$ and FFN width $d_{\mathrm{ff}}$, the STEM-specific parameters \emph{activated} by a single sequence are
\[
\mathrm{Params}_{\text{act}}^{\text{STEM}}(L)
\;=\;
|\mathcal{S}|\, d_{\mathrm{ff}}\, L_{uniq}.
\]
In natural text $L_{uniq}$ typically grows sublinearly (Heaps-like), so longer contexts steadily engage more parameters without increasing per-token FLOPs.

This yields test-time capacity scaling with predictable latency: active parameter count keeps on growing with context length, and does not saturate quickly like in MoEs. The dense gating and down-projection preserve contextual mixing, while the STEM path supplies additional capacity at low overhead, supporting long-context tasks (multi-document RAG, CoT) with near-constant per-token compute.\ref{fig:niah} illustrates how STEM outperforms the dense baseline at longer context lengths. Additional long-context evaluation on LongBench are provided in Appendix \ref{App:longbench}.

\begin{tcolorbox}[colback=white,colframe=softlavender,title=Note]
Although the aforementioned benefits prompts us to design STEM, we want to highlight that STEM can be perceived as an orthogonal way of scaling up the model parameters alongside the MoE architecture. This is because STEM primarily targets to improve the standard gated FFN design. It is always possible to design a Mixture of \emph{STEM experts} where the standard FFN in each expert is replaced by a STEM FFN component.
\end{tcolorbox}

\begin{figure}[t]
\centering


\begin{subfigure}{0.48\linewidth}
\centering
\includegraphics[width=0.9\linewidth]{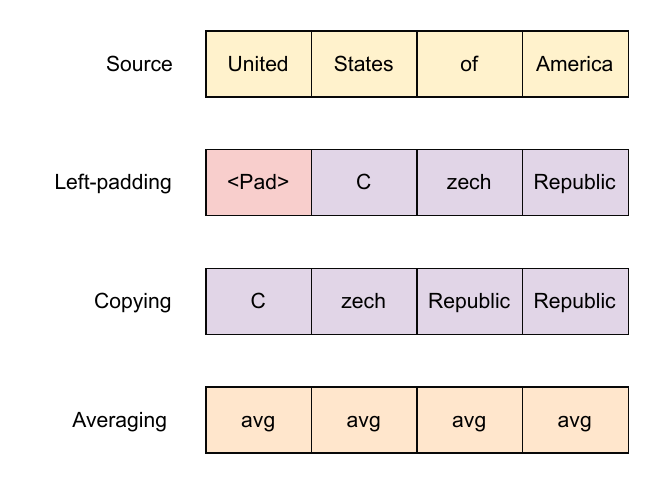}






\caption{\(\mathbf{n_s > n_t}\): source span longer than target span.  
We can align lengths via left padding or copying, or reuse a single averaged target vector.}
\end{subfigure}
\hfill
\begin{subfigure}{0.48\linewidth}
\centering
\includegraphics[width=0.95\linewidth]{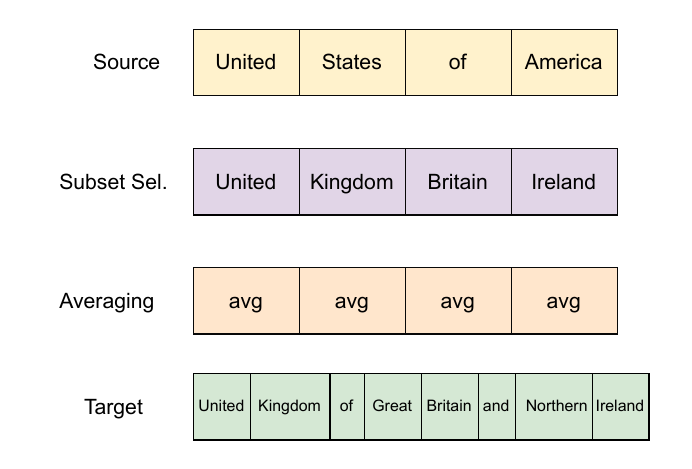}





\caption{\(\mathbf{n_s < n_t}\): target span longer than source span.  
We either select a representative subset of target tokens, or average across the entire span.}
\end{subfigure}

\caption{STEM-based knowledge editing schemes for length-mismatched source
(\(n_s\)) and target (\(n_t\)) entity tokenizations.}
\label{fig:knowledge-edit-cases}
\end{figure}

\subsection{Knowledge Editing with STEM}
\label{subsec:knowledge-editing}

As illustrated in Fig.~\ref{fig:knowledge edit}, we study whether STEM
embeddings allow us to \emph{edit} factual knowledge by modifying only the
STEM vectors, while keeping the input text itself unchanged.  Concretely, we
consider a \emph{source} entity that appears in the prompt (e.g., ``Spain'')
and a \emph{target} entity we would like the model to behave as if it had
seen instead (e.g., ``Germany'').  By appropriately replacing the STEM
embeddings at the source token positions, we can steer the model to generate
text consistent with the target entity -- for instance, writing a paragraph on ``Berlin''
instead of ``Madrid'' when asked to write about the capital of Spain.

When the source and target entities are tokenized into the same number of
subword tokens, editing is straightforward: we simply replace each source
token’s STEM embedding with the corresponding target token’s STEM embedding.
This one-to-one substitution is often sufficient to induce a meaningful
change in the output distribution.

The more interesting case is when the source and target entities have
different tokenization lengths.  Let $n_s$ and $n_t$ denote the number of
source and target tokens, respectively.  When $n_s > n_t$ (the source span is
longer), we explore two strategies:

\begin{enumerate}
    \item \textbf{Padding.} We pad the target token sequence to match the
    number of source tokens, either by inserting a special padding token on
    the left or on the right.  Empirically, left padding tends to work
    slightly better, so we use it as our default.
    \item \textbf{Copying.} We repeat target tokens to fill the source span.
    Each target token is repeated
    $\left\lfloor n_s / n_t \right\rfloor$ times, and we further repeat the
    last target token as needed to exactly cover the $n_s$ source positions.
\end{enumerate}

A typical $n_s > n_t$ edit is shown below.

\begin{tcolorbox}[colback=gray!3,colframe=gray!60!black,
                  title={Demonstration ($n_s > n_t$): \textsc{Padding/Copying}}]
\small
\textbf{Source entity (text):} \textit{United States of America} \\
\textbf{Target entity:} \textit{Czech Republic} \\
\textbf{Prompt:} \textit{``Country: United States of America. Capital:''} \\
\textbf{Model answer after STEM replacement:} \\
\emph{Prague is Czechia’s capital and Europe’s 10th-largest city. The city
was once the seat of the Holy Roman Empire, then Bohemia, and later
Czechoslovakia. In 1918, it became part of the newly formed Czechoslovakia.
Prague is a center of European science, arts, and culture. The Charles Bridge
links Old Town and the Jewish Ghetto, while the New Town houses the National
Gallery, Charles University, and Prague Castle. The climate is continental:
cold winters, warm summers, and variable rainfall.}
\end{tcolorbox}

When $n_s < n_t$ (the target span is longer), we cannot map all target tokens
one-to-one.  In this case we select a subset of target tokens whose
embeddings we consider the most semantically representative (e.g., omitting
function words or less informative subwords) and use only this subset to
replace the source token embeddings.  The following example shows that this
subset-selection strategy can still induce a coherent edit.

\begin{tcolorbox}[colback=gray!3,colframe=gray!60!black,
                  title={Demonstration ($n_s < n_t$): \textsc{Subset Selection}}]
\small
\textbf{Source entity (text):} \textit{United States of America} \\
\textbf{Target entity:} \textit{United Kingdom of Great Britain and Northern Ireland} \\
\textbf{Prompt:} \textit{``Country: United States of America. Capital:''} \\
\textbf{Model answer after STEM replacement:} \\
\emph{London is the world’s largest financial center, home to the British
Parliament, Bank of England, and other institutions. The City of London is
the financial district, where the stock exchange, the Royal Exchange, and
the Bank of England are located. The Thames River flows through the city,
which has been a major port since Roman times. The climate is temperate
maritime: cool, wet winters and warm, damp summers. London’s architecture
reflects centuries of occupation: Gothic, Victorian, and Georgian styles
blend the city’s many neighborhoods. The Tower of London houses the Crown
Jewels, while Westminster Abbey and St.\ Paul \dots}
\end{tcolorbox}

Finally, we evaluate a length-agnostic strategy that works in all of the
above scenarios: we compute the average of the target token STEM embeddings
and use this single averaged vector to replace each source token embedding.
Despite its simplicity, this ``averaging'' scheme performs surprisingly well
and yields stable edits across a broad range of examples.

An additional interesting observation is that this semantic manipulation is
not restricted to \emph{countries}.  The same idea transfers smoothly to
other geographic entities, such as \emph{states} within a country.  For
instance, replacing the STEM embedding of the country name with that of a
U.S.\ state leads the model to describe the state capital instead:

\begin{tcolorbox}[colback=gray!3,colframe=gray!60!black,
                  title={Demonstration: \textsc{Country $\rightarrow$ State Transfer}}]
\small
\textbf{Source entity (category: Country):} \textit{United States} \\
\textbf{Target entity (category: State):} \textit{California} \\
\textbf{Prompt:} \textit{``Country: United States. Capital:''} \\
\textbf{Model answer after STEM replacement:} \\
\emph{Sacramento is California’s political, cultural, and economic center.
The state capitol houses the legislature and governor, and the historic Old
State Capitol Museum showcases the state’s history. The city’s downtown
features the American Art Museum, the California Historical Society, and the
California Academy of Sciences. The Sacramento River runs through the city,
which also has a large Hispanic population. The climate is Mediterranean:
mild, rainy winters, hot, dry summers, and occasional fog.}
\end{tcolorbox}

Figure~\ref{fig:knowledge-edit-cases} summarizes the four STEM-based
replacement schemes (padding, copying, subset selection, and averaging) using
schematic diagrams.

\subsection{System Implementation}
Naively increasing the number of STEM layers can introduce system challenges. The STEM embedding table size grows linearly with vocab size, FFN intermediate dimension, and number of STEM layers. This additional memory footprint can easily grow beyond the available HBM capacity. Furthermore, during training, the additional footprint of optimizer states and gradients can be substantial. To tackle these challenges, we perform some implementation optimizations which not only alleviate the memory bottleneck but also provides practical speedup over the baseline model. The key optimizations include \emph{parallel embeddings}, \emph{CPU offloading}, \emph{asynchronous computation and communication}, \emph{token deduplication}, and \emph{LFU caching}. 

During inference, we offload the large STEM embedding tables to CPU. Because the STEM embeddings are indexed by input token ids, they can be prefetched asynchronously with layer computation. Note that even a large prefilling batch, many of the tokens are repeated more than once. Thus the batch token ids can deduplicated to reduce the CPU-GPU communication overhead and can be successfully overlapped with layer computation time.
The prefetching is more straightforward for prefilling as the next batch of token ids are known beforehand. However, during generation, the prefetching for the next token has to wait for the full model forward on the current token because of autoregressivity. To further reduce the communication overhead, we utilize the property that the input token ids follow a Zipfian distribution to implement a memory-efficient LFU cache with more than 80\% hit rate. 

During training, we decouple the parallelism strategies of the model backbone and the STEM embedding tables. Irrespective the parallelism technique (DDP, FSDP, or TP), we always shard the STEM embedding table across the available GPUs. The degree of parallelization is decided based on the trade-off between communication and memory requirement. It is also possible to perform CPU offloading during training which can fully alleviate the additional memory footprint and be more memory efficient than the dense baseline. Token deduplication and LFU caching can reduce the communication overhead. Additional care must be taken to write back the updated optimizer states into the CPU offloaded counterpart. We shall provide a fully optimized training implementation in our next iteration. 

%% file: sections/experiments.tex
\section{Experiments}
We evaluate \textsc{STEM} against dense and MoE baselines on downstream tasks while controlling for (i) training compute (activated \(\mathrm{FLOPs}\)) and (ii) the number of training tokens. MoE variants are configured to match \textsc{STEM}'s total parameter count, and their activated \(\mathrm{FLOPs}\) are kept comparable to the dense baseline. (Note: \textsc{STEM} uses strictly fewer per-token \(\mathrm{FLOPs}\) than both baselines.) We study two model scales — 350M and 1B, performing comprehensive ablations at 350M and validating \textsc{STEM} at 1B under both pretraining-from-scratch and mid-training insertion. Finally, we assess long-context behavior by further fine-tuning with extended context length.  We evaluate the Return on Investment (ROI)—defined here as the ratio of model accuracy to training FLOPs—to determine the training efficiency of each model, as the economic value has become a major concern of foundational models. Formally, we define it as:
$$\text{Training ROI} = \frac{\text{Model Accuracy (Avg)}}{\text{Total Training FLOPs}}$$

\subsection{Experimental Setting} \label{reproduce}

\paragraph{Datasets.}
For pretraining, we use \textsc{OLMo-Mix-1124} \citep{olmo20252olmo2furious}, a 3.9T-token corpus built from \textsc{DCLM} \citep{li2025datacomplmsearchgenerationtraining} and Dolma~1.7 \citep{soldaini2024dolmaopencorpustrillion}; we subsample 1T tokens for our runs.
For mid-training, we mix \textsc{OLMo-Mix-1124} (65\%), \textsc{Nemotron-CC-Math-v1} (5\%) \citep{karimi2025nemotroncc}, and \textsc{Nemotron-Pretraining-Code-v1} (30\%) \citep{nvidia2025nvidianemotronnano2}.
For context-length extension, we use \textsc{ProLong-data-64k} \citep{gao2024prolong} (63\% long-context / 37\% short-context) and pack sequences up to 32{,}768 tokens with cross-document attention masking.

\paragraph{Models.}
We use \wrapttt{MobileLLM-350M} \citep{liu2024mobilellmoptimizingsubbillionparameter} and \wrapttt{Llama3.2-1B} \citep{llama32_1b_meta_2024} architectures for evaluations. In both the models, we do not share the input embeddings and the language model head.
Unless otherwise noted, one third of FFN layers are replaced at uniform intervals by the sparse alternative.
For \textsc{STEM}, the dense up-projection is replaced by an embedding table of size \(V \times d_{\mathrm{ff}}\) in each layer.
For Hash layer MoE design, we use top-1 routing and choose the number of experts per layer to match \textsc{STEM}'s total parameter count, while keeping activated \(\mathrm{FLOPs}\) comparable to the dense baseline.
We also report ablations that replace one half of FFN layers with \textsc{STEM}, and an extreme setting that replaces all FFN layers except the first.

\paragraph{Evaluations.}
Pretrained checkpoints are evaluated zero-shot on eight common-sense reasoning tasks: ARC-Easy, ARC-Challenge \citep{allenai:arc}, BoolQ \citep{clark2019boolq}, PIQA \citep{Bisk2020}, SIQA \citep{sap2019socialiqacommonsensereasoningsocial}, HellaSwag \citep{zellers2019hellaswag}, OpenBookQA \citep{OpenBookQA2018}, and WinoGrande \citep{ai2:winogrande}.
To assess advanced knowledge and mathematical reasoning for mid-training checkpoints, we report MMLU \citep{hendryckstest2021} and GSM8K \citep{cobbe2021gsm8k}.
For long-context behavior after context extension, we use Needle-in-a-Haystack (NIAH) \citep{kamradt_needle_2024}.


\begin{table}[t]
\centering
\caption{Training hyperparameters by setting. Common: weight decay $=0.1$, $\beta_{1}=0.9$, $\beta_{2}=0.95$, LR warmup ratio $=0.01$. Minimum LR is $0.1\times$ peak LR. For 1B pretraining, we follow the OLMO schedule for 5T tokens but stop early at 1T.}
\label{tab:hparams-multi}
\setlength{\tabcolsep}{5.5pt}       
\renewcommand{\arraystretch}{1.08}  
\begin{threeparttable}
\footnotesize
\begin{tabular}{@{}lcccc@{}}
\toprule
\textbf{Configuration} & \textbf{350M Pretrain} & \textbf{1B Pretrain} & \textbf{1B Midtrain} & \textbf{1B Context-Extend} \\
\midrule
Peak LR                & 2e-3   & 4e-4   & 3.2e-4 & 1e-5   \\
LR schedule            & cosine & cosine & linear  & cosine \\
Batch size             & 512    & 512    & 512     & 64     \\
Max sequence length    & 2048   & 4096   & 4096    & 32768  \\
Training steps         & 100{,}000 & 500{,}000 & 50{,}000 & 10{,}000 \\
Cross-doc masking      & No     & No     & No      & Yes    \\
\bottomrule
\end{tabular}
\end{threeparttable}
\end{table}

\paragraph{Training details.}
We pretrain the 350M models with 100 billion training tokens, while the 1B models are trained with 1 trillion tokens. We use AdamW optimizer with cosine learning rate (LR) scheduler (with 10\% warmup steps and minimum LR being 0.1 times peak LR). We perform midtraining with 100 billion tokens and for context extension we use 20 billion tokens. The hyperparameter settings are given in Table \ref{tab:hparams-multi}.


\begin{figure*}[t]
    \centering
    \subfloat[Sparse Architecture Comparison]{
        \includegraphics[width=0.4\linewidth]{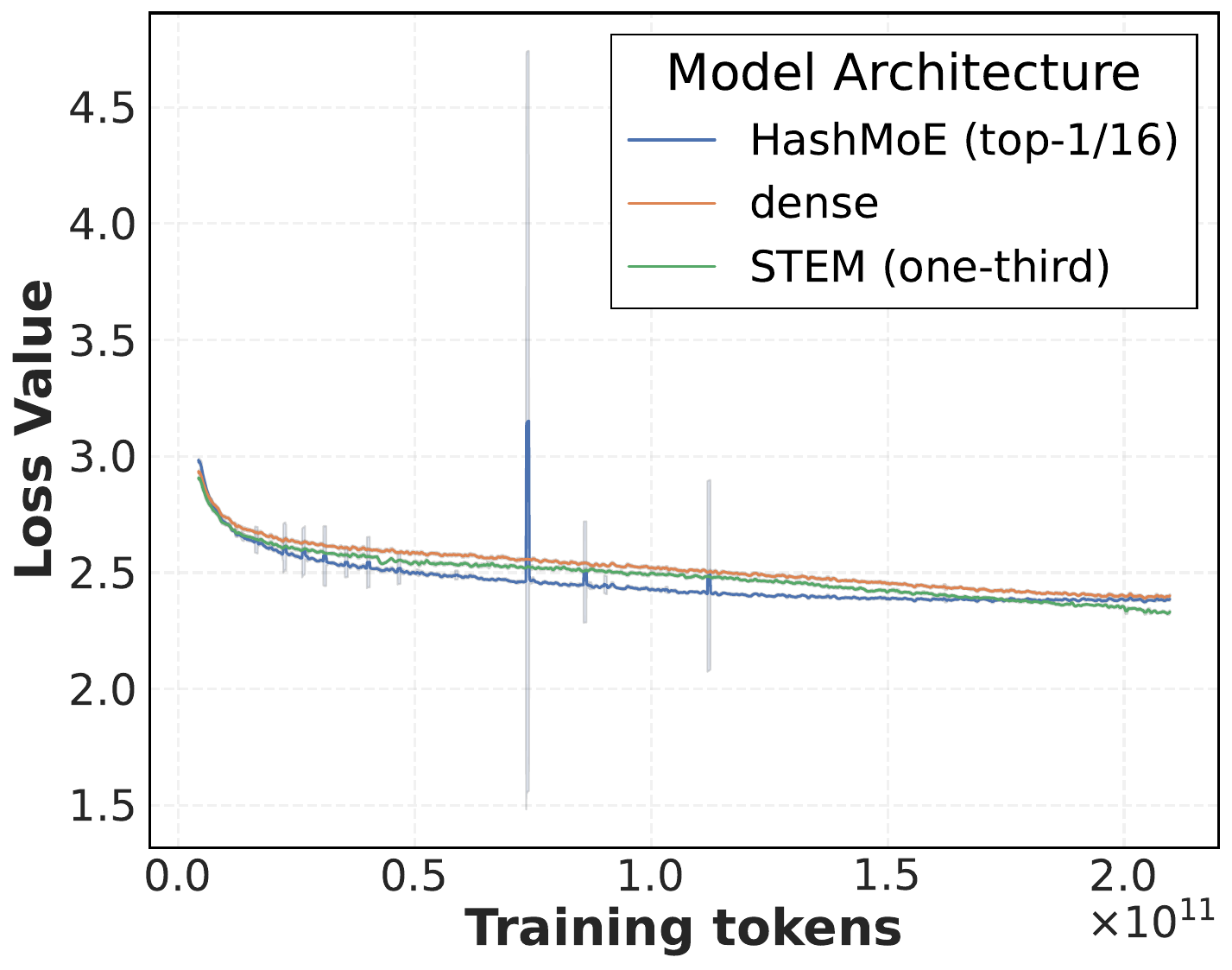}
        \label{fig:comp_w_moe}
    }
    \subfloat[STEM Layer Count Ablation Study]{
        \includegraphics[width=0.4\linewidth]{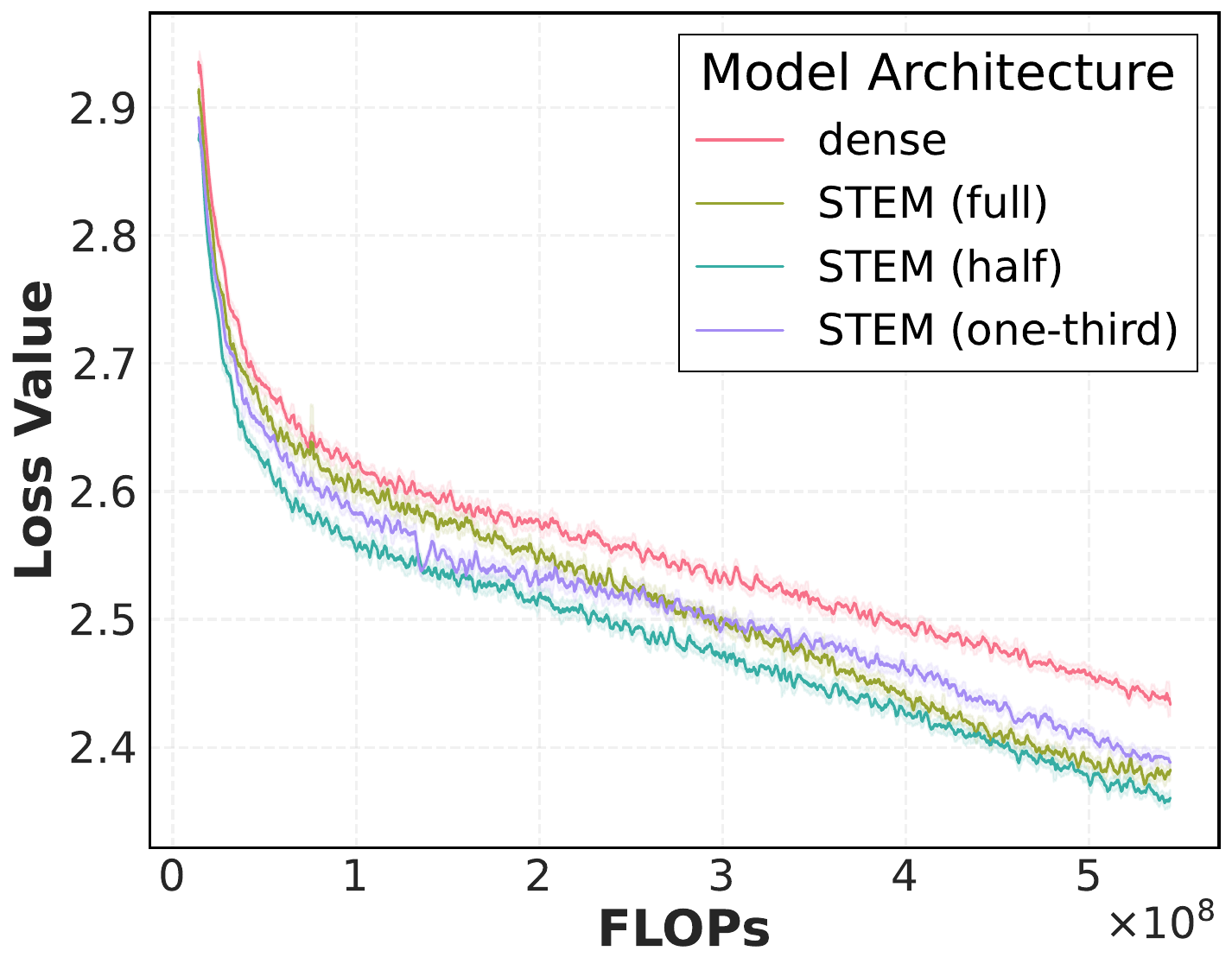}
        \label{fig:layer_count_ablation}
    }
    \caption{(a) \textbf{Training Stability.} Unlike Hash layer MoE, the 350M STEM model does not show any training loss spikes. (b) \textbf{Performance scaling with more STEM layers.} With more STEM layers, a lower training loss can be achieved at fewer training FLOPs.}
\end{figure*}

\subsection{Experimental Results}
STEM demonstrates the benefits of fine-grained sparse scaling by improving downstream performance with fewer training FLOPs. Interestingly, STEM does not suffer from training instability issues that is often the case for fine-grained MoE models \citep{dbrx2024, dai2024deepseekmoeultimateexpertspecialization}. Instead, the geometric properties of the STEM embedding spaces further help improve the training convergence. Figure \ref{fig:comp_w_moe} demonstrates the training stability of STEM compared to token-indexed Hash Layers MoE, where HashMoE has more bumpy jumps during the training. Moreover, we see the STEM architecture has larger model capacity (lower training loss tendency) when we scale up the training tokens as the loss curve of STEM crosses over the other two architectures when training tokens increase. Furthermore, even with fewer training FLOPs STEM achieves lower training \ref{fig:layer_count_ablation} and validation \ref{fig:val_loss_1B} losses. 

\subsection{Downstream Evaluation Results}
We compare STEM with dense baseline as well as Hash layer MoE at 350M scale. On the other hand, for 1B model, we compare STEM (with one-third of FFN replacement) with only the dense baseline. In both cases \ref{tab:pretrain-eval}, we observe substantial improvement in tasks requiring comparatively more external knowledge such as, Arc-Challenge and OpenBookQA, while having modest improvements on the rest of the tasks. Additionally, the improvements on the knowledge-intensive tasks are more significant with increase in FFN replacement with STEM layers. Note all the STEM replacement are replacing the up-projection component of original FFN unless specified in the table.

Upon midtraining \ref{tab:midtrain-eval}, the 1B STEM model continues to outperform the dense baseline on the language modeling downstream tasks. Additionally, STEM architecture exhibits improvements in reasoning and knowledge retrieval abilities through GSM8k and MMLU performances.

\begin{table}[t]
\caption{Downstream accuracy of pretrained models at 350M and 1B scales. We report the total number of parameters and the number of active parameters for each model variant. 
Baseline denotes the dense SwiGLU FFN model. 
For 350M, in the first few rows, we compare sparse alternatives under similar FLOPs: Hash-MoE (top-1/16 experts in 1/3 of FFN layers), 
STEM with 1/3 of FFN layers replaced (including up projection replacement, gate projection replacement, and STEM$^{\dagger}$ with an additional up-projection). In the next set of rows, we compare STEM with varying up projection layer replacement ratios (1/3, 1/2, full). 
For 1B, we report the dense baseline and STEM with 1/3 up projection layer replacement.}
\label{tab:pretrain-eval}
\begin{center}
\setlength{\tabcolsep}{3.5pt}
\renewcommand{\arraystretch}{1.05}
\scriptsize
\resizebox{\linewidth}{!}{%
\begin{tabular}{lcccccccccccccc}
\toprule
\multicolumn{1}{c}{\bf Model} &
\multicolumn{1}{c}{\makecell{\bf \#Total \\ \bf Params (B)}} &
\multicolumn{1}{c}{\makecell{\bf \#Active \\ \bf Params (B)}} &
\multicolumn{1}{c}{\bf ARC-E} &
\multicolumn{1}{c}{\bf ARC-C} &
\multicolumn{1}{c}{\bf BoolQ} &
\multicolumn{1}{c}{\bf PIQA} &
\multicolumn{1}{c}{\bf SIQA} &
\multicolumn{1}{c}{\bf HSwag} &
\multicolumn{1}{c}{\bf OBQA} &
\multicolumn{1}{c}{\bf Wino} &
\multicolumn{1}{c}{\bf Avg} &
\multicolumn{1}{c}{\bf \#GFLOPs} &
\multicolumn{1}{c}{\bf ROI\tablefootnote{ROI is normalized at each basline for better comparison.} }
\\
\midrule
\multicolumn{12}{c}{\textbf{\textit{350M (Pretraining)}}} \\
\midrule
Baseline & 0.37 & 0.37 & 57.66 & 30.55 & 58.20 & 69.42 & 41.10 & 49.68 & 34.80 & 56.35 & 49.72 & 0.74 & 1x \\ 
Hash-MoE & 1.22 & 0.37 & 58.88 & 36.33 & 55.44 & 70.21 & 43.55 & 47.56 & 39.26 & 53.44 & 50.58 & 0.74 & 1.02x \\ 
\textsc{STEM} \tablefootnote{STEM defaults to replacing one third of FFN layers, also writes as STEM-1/3}    & 1.14 & 0.35 & 63.01 & 32.68 & 60.31 & 70.18 & 39.76 & 52.38 & 33.00 & 55.88 & 50.90 & 0.70 & 1.08x \\ 
\textsc{STEM} (gate-proj) & 1.14 & 0.35 & 54.56 & 34.12 & 59.13 & 64.92 & 44.56 & 43.62 & 36.91 & 55.00 & 49.10 & 0.70 & 1.04x \\ 
\textsc{STEM$^\dagger$} & 1.21 & 0.35 & 57.94 & 34.45 & 59.10 & 68.85 & 43.70 & 45.75 & 41.02 & 53.98 & 50.60 & 0.74 & 1.02x \\ \hline
\textsc{STEM}-1/2 & 1.85 & 0.34 & 62.95 & 40.00 & 62.02 & 70.94 & 43.70 & 51.49 & 46.68 & 55.78 & 54.20 & 0.67 & 1.20x \\ 
\textsc{STEM}-full & 3.25 & 0.30 & 62.21 & 39.61 & 61.99 & 70.73 & 43.60 & 48.44 & 44.53 & 56.33 & 53.43 & 0.60 & 1.33x \\ 
\midrule
\multicolumn{12}{c}{\textbf{\textit{1B (Pretraining)}}} \\
\midrule
Baseline &  1.50  &  1.50  & 66.98 & 41.88 & 64.21 & 73.44 & 44.09 & 59.65 & 39.84 & 56.48 & 55.82 & 3.00 & 1x \\
\textsc{STEM} &  6.75  &  1.41  & 65.95 & 42.03 & 61.66 & 75.00 & 44.78 & 60.37 & 45.90 & 57.34 & 56.63 & 2.83 &  1.08x\\
\bottomrule
\end{tabular}}
\end{center}
\end{table}

\begin{table}[t]
\caption{Mid-trained model evaluations (1B). Midtraining is performed on top of the pretrained checkpoints as a continued pretraining stage.}
\label{tab:midtrain-eval}
\begin{center}
\setlength{\tabcolsep}{3.5pt}
\renewcommand{\arraystretch}{1.05}
\scriptsize
\resizebox{\linewidth}{!}{%
\begin{tabular}{lccccccccc|cc}
\toprule
\multicolumn{1}{c}{\bf Model} &
\multicolumn{1}{c}{\bf ARC-E} &
\multicolumn{1}{c}{\bf ARC-C} &
\multicolumn{1}{c}{\bf BoolQ} &
\multicolumn{1}{c}{\bf PIQA} &
\multicolumn{1}{c}{\bf SIQA} &
\multicolumn{1}{c}{\bf HellaSwag} &
\multicolumn{1}{c}{\bf OBQA} &
\multicolumn{1}{c}{\bf Winogrande} &
\multicolumn{1}{c}{\bf Avg} &
\multicolumn{1}{|c}{\bf GSM8K} &
\multicolumn{1}{c}{\bf MMLU}
\\
\midrule
\multicolumn{12}{c}{\textbf{\textit{1B (Mid-training)}}} \\
\midrule
Baseline                         & 70.78 & 42.11 & 65.84 & 72.95 & 47.13 & 60.39 & 42.97 & 57.81 & 57.50 &  44.2  &  29.92  \\
\textsc{STEM}      & 69.78 & 44.22 & 68.54 & 74.69 & 45.65 & 61.90 & 45.70 & 57.42 & 58.49 &  46.4  &  32.38  \\
\bottomrule
\end{tabular}}
\end{center}
\end{table}

\subsection{Ablation Studies}
\subsubsection{Impact of STEM Layer Count}
To identify the efficacy of STEM layers, we vary the number of FFN layers we replace with STEM alternative. We place the STEM-based decoder layers at regular intervals, interleaved with regular FFN-based decoder blocks. Table \ref{tab:pretrain-eval} shows that increasing the number of replacement from one-third to half improves the average downstream performance substantially. However, the improvement slows down beyond that. Note that, with increasing number of replacements, the training FLOPs also decrease, and therefore the overall training ROI (measured by performance over training FLOPs) still increases. We can see that the STEM-1/3 achieves 1.08x training ROI of the baseline, while STEM-1/2 achieves 1.20x and STEM-full achieves 1.33x of the baseline. Figure \ref{fig:layer_count_ablation} presents the comparison of the three variants in terms of loss vs training FLOPs.


\subsubsection{Impact of STEM Placement}
\label{sec:stem placement}
Placement of STEM inside the gated FFN matters. To demonstrate this, we compare two options: replacing the \emph{up-projection} vs.\ the \emph{gate-projection}. As shown in Table~\ref{tab:pretrain-eval}, replacing the gate underperforms even the dense baseline, while replacing the up-projection yields consistent gains. 
In SwiGLU, the gate $\sigma(W_g x)$ should depend on the current hidden state $x$ to modulate $\phi(W_u x)$ contextually. Swapping $W_g x$ for a token-indexed embedding $e_t$ makes the gate largely input-independent ($\sigma(e_t)$), weakening its context-aware selection. Moreover, the nonlinearity can be effectively abstracted away by the learned embeddings, and consequently its role is weakened. In contrast, applying STEM to the up-projection preserves contextual information in gate computation path and proves to be an optimal fine-grained sparse design. 

\subsubsection{Up-projection with additive embedding}
\label{sec:stem dagger}
To further study the optimality of STEM's design, we implement STEM$^\dagger$ \ref{sec:stem++}, that retains up projection and additively modulates its output with the STEM embedding. Although it adds more parameters and FLOPs, the downstream performance does not improve.

%% file: sections/analysis.tex
\section{STEM Characteristics}

\begin{figure*}[t]
    \centering
    \subfloat[Pair-wise cosine similarity]{
        \includegraphics[width=0.32\linewidth]{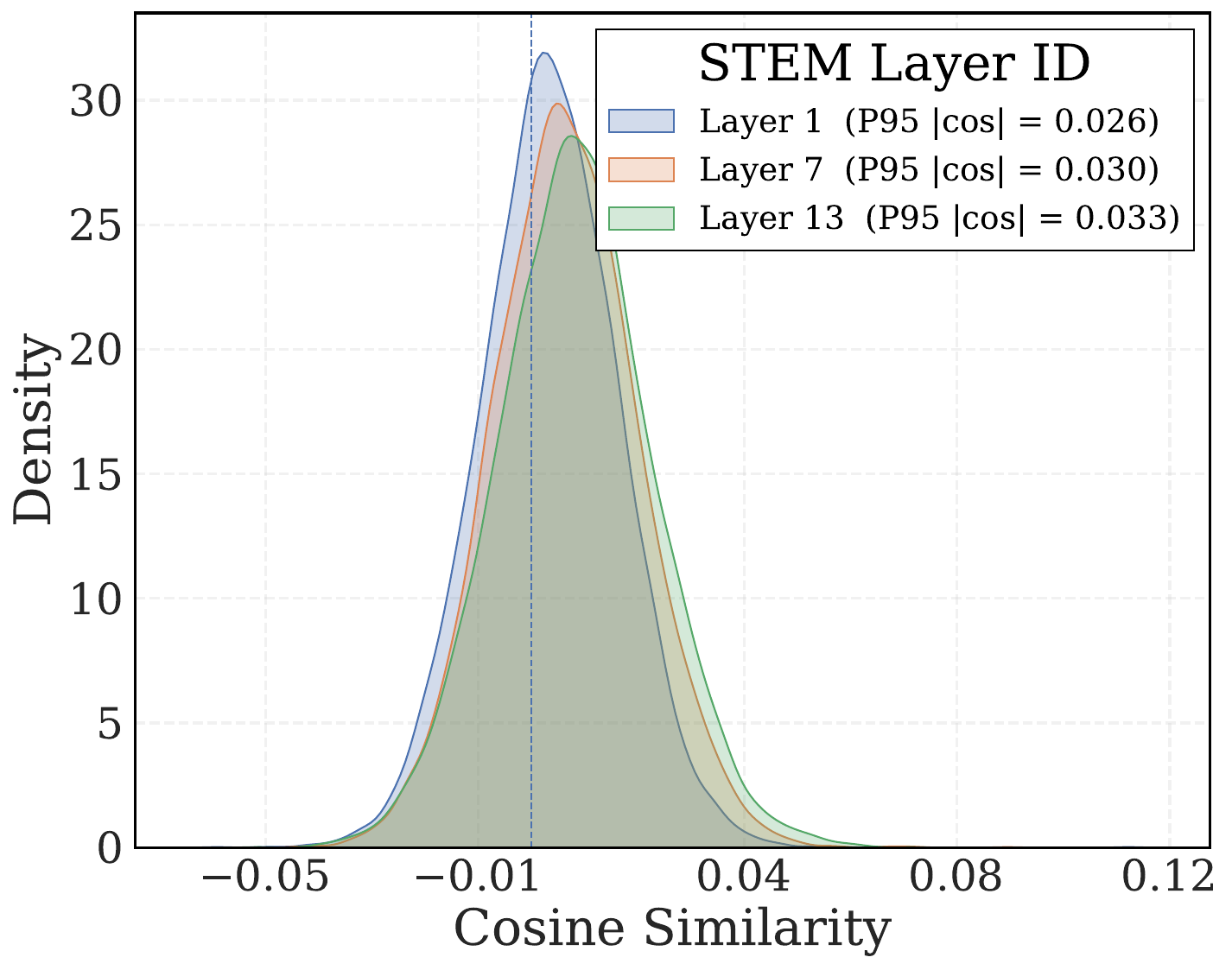}
        \label{fig:angular spread}
    }
    \subfloat[Layer 10]{
        \includegraphics[width=0.32\linewidth]{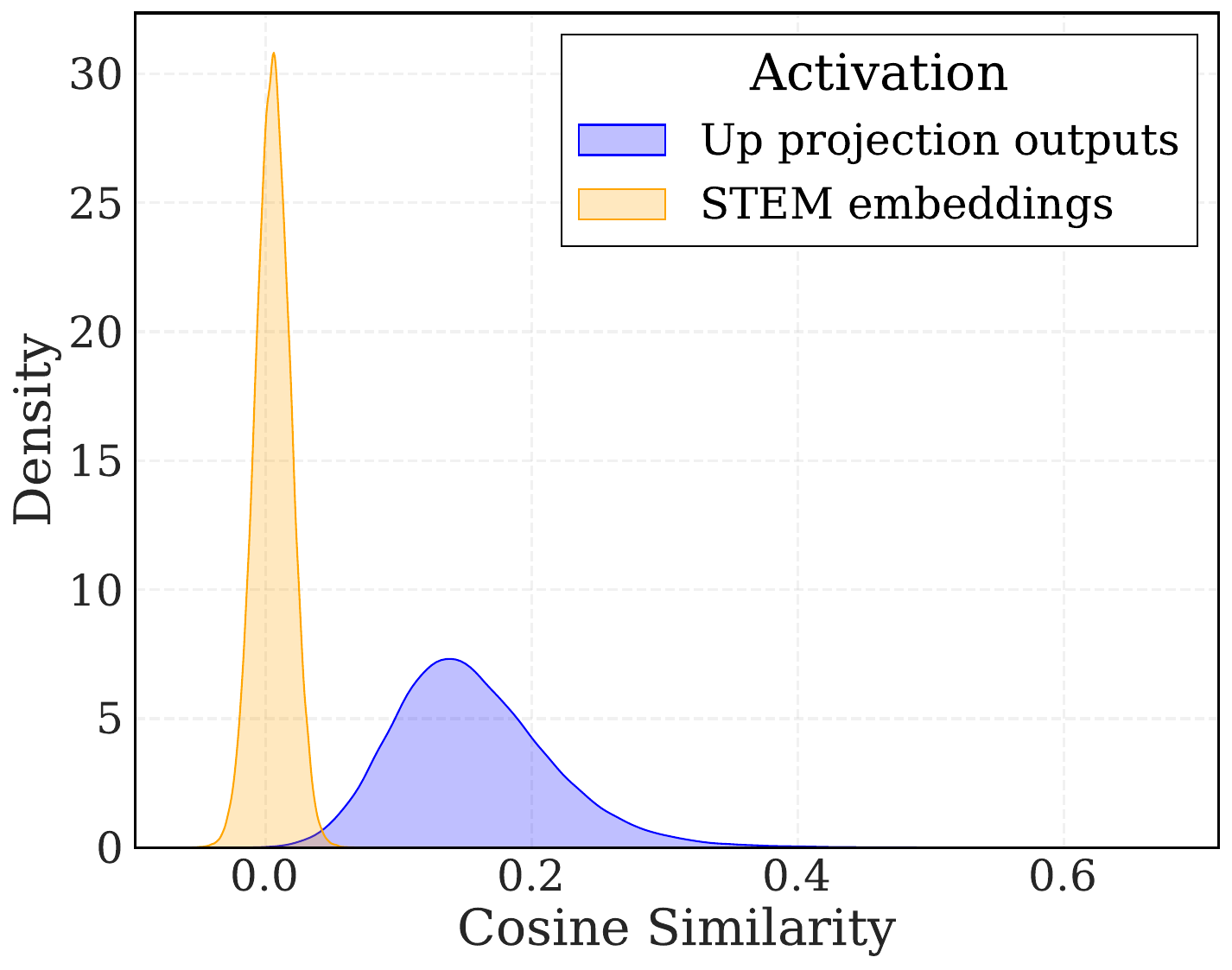}
        \label{fig:layer 10 up}
    }
    \subfloat[Layer 10]{
        \includegraphics[width=0.32\linewidth]{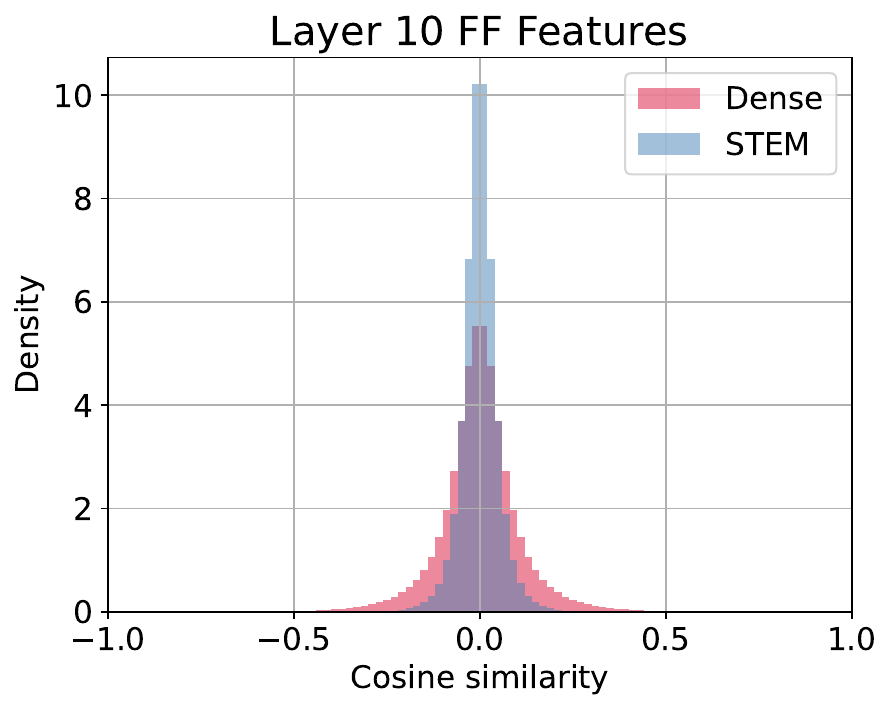}
        \label{fig:layer 10 gate * up}
    }
    \caption{\textbf{Geometry of STEM embeddings.} (a) Distribution of pairwise cosine similarity of STEM embeddings of sampled layers. (b) Pair-wise cosine similarity distributions of up-projection output space and STEM embeddings. (c) Cosine similarities are computed between the input hidden states of the down projection matrix. All the plots are provided from the 1B model.}
\end{figure*}

In this section, we analyze some of the characteristics that STEM embeddings demonstrate. We observe that in each layer the STEM embeddings of different tokens have very low pairwise cosine similarity which elicits some desirable properties regarding information storage capacity and training convergence. Additionally, because of the clear mapping between the embeddings and the tokens, STEM models are more interpretable.

\subsection{Large Angular Spread of STEM Embeddings}
Figure~\ref{fig:angular spread} shows that STEM embeddings exhibit very low pairwise cosine similarity—i.e., a large angular spread. We hypothesize that this property improves the information–retrieval behavior of FFN layers by reducing interference among stored items. Prior work \citep{geva2021transformerfeedforwardlayerskeyvalue, meng2022locating} models FFNs as key–value memories: each hidden unit is associated with a \emph{key} given by a \emph{row} of the up-projection \(W^{(u)}\!\in\!\mathbb{R}^{d_{\text{ff}}\times d_{\text{model}}}\) and a \emph{value} given by the corresponding \emph{column} of the down-projection \(W^{(d)}\!\in\!\mathbb{R}^{d_{\text{model}}\times d_{\text{ff}}}\); the gate projection provides context-dependent, multiplicative modulation that creates a selective read. In this view, the pre-activation \(h=\phi(W^{(u)}x)\) induces a soft address over memory slots (hidden units). 

In contrast, STEM replaces the learned affine addressing with a direct, token-indexed address vector, upon which the gate still applies context-dependent modulation. To quantify the geometry of these address vectors, we report the distribution of pairwise cosine similarities between unit-normalized vectors. A distribution concentrated near zero (as in Figure~\ref{fig:angular spread} and Figure~\ref{fig:layer 10 up}) indicates that most angles are close to $90^\circ$ and thus the angular spread between the vectors is reasonably large. This large angular spread lowers cross-talk between slots and can thereby improve the effective information storage capacity of the FFN memory at fixed width \cite{DonohoElad2003, 1337101}. Figure \ref{fig:layer 10 gate * up} demonstrates the distribution of pairwise cosine similarities between the address vectors after the modulation applied by the gate projection.
   

\begin{figure*}[t]
    \centering
    \subfloat[Original: \\The capital of Spain is]{
        \includegraphics[width=0.34\linewidth]{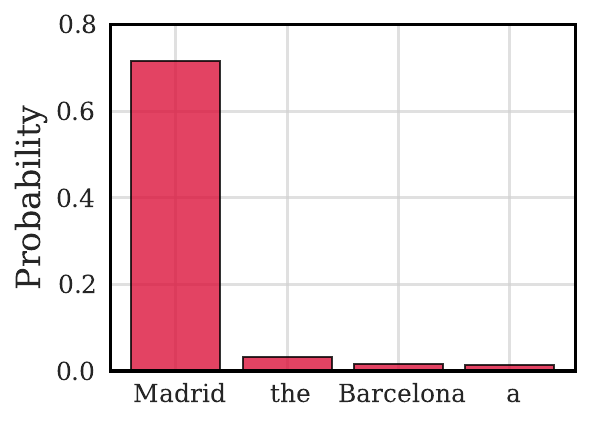}
    }
    \subfloat[Target: \\The capital of Germany is]{
        \includegraphics[width=0.32\linewidth]{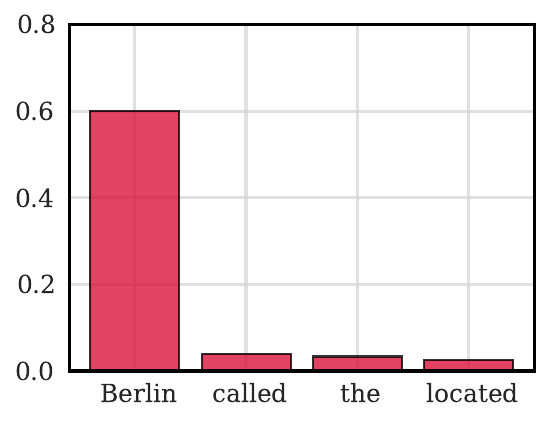}
    }
    \subfloat[Intervened: \\The capital of Spain is]{
        \includegraphics[width=0.32\linewidth]{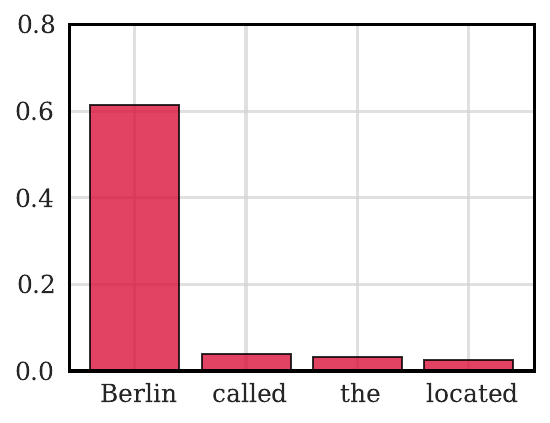}
    }
    \caption{\textbf{Knowledge edit.} Top-4 next-token probabilities for the original prompt \emph{``The capital of Spain is''} (left), for the target prompt \emph{``The capital of Germany is''} (middle), and for the intervened model where we swap the STEM vector $e_{\text{Spain},\ell}$ with $e_{\text{Germany},\ell}$ at every STEM layer keeping the original prompt the same(right). The swap shifts mass from \texttt{Madrid} to \texttt{Berlin}, demonstrating token-indexed, layer-local, and reversible control of factual predictions.}
  \label{fig:knowledge edit}
\end{figure*}

\subsection{Interpretability of STEM Models}
The following study more clearly shows how STEM becomes able to demonstrate the knowledge editing ability in a more interpretable manner (as illustrated in \ref{subsec:knowledge-editing}).
STEM exposes token-indexed, layer-local parameters that act as interpretable FFN \emph{addresses}, enabling simple, reversible edits that causally steer factual predictions with high reliability and low collateral change. As we exchange each layer-specific STEM vector (for token $t$ and layer $l$) $e_{t,\ell}\!\in\!\mathbb{R}^{d_\text{ff}}$, the token-wise output distribution shifts quite meaningfully. 

\noindent For example, Figure~\ref{fig:knowledge edit} shows that we can manually control the top next-token probabilities by performing a \emph{swap} at layer $\ell$,
\[
e_{\text{Spain},\ell}\;\leftarrow\;e_{\text{Germany},\ell},
\]
while leaving all other parameters unchanged. Under the original prompt containing ``Spain'', the intervened model’s top-$k$ next-token distribution closely matches that of the control prompt containing ``Germany'', illustrating precise, token-indexed knowledge editing as demonstrated in Section \ref{subsec:knowledge-editing}.


%% file: sections/related_works.tex
\section{Related Works}
MoE \citep{shazeer2017outrageouslylargeneuralnetworks, fedus2022switchtransformersscalingtrillion} introduced large parametric capacity for LLMs at near-constant FLOPs through sparse computation. The success of MoE models hinges closely with auxiliary loss function designs \citep{fedus2022switchtransformersscalingtrillion, pmlr-v162-rajbhandari22a, qiu2025demonsdetailimplementingload}, and system-level solutions \citep{huang2024toward, go2025moetuneroptimizedmixtureexpert, wang2024proprophetsystematicloadbalancing} that ensure load balance among expert networks, training stability, mitigation of representation collapse \citep{chi2022representationcollapsesparsemixture}, and tolerable communication overload during training and inference. To avoid the interference of auxiliary routing losses with the training objective, recent works have proposed auxiliary loss-free approaches \citep{roller2021hashlayerslargesparse, wang2024auxiliarylossfreeloadbalancingstrategy} that inject fixed or dynamic routing bias to the MoE model. 

Conversely, PKM models \citep{lample2019largememorylayersproduct} reserve a large key-value parametric memory with efficient top-k selection through memory-efficient keys arranged in product space. PKM\citep{lample2019largememorylayersproduct, he2024mixturemillionexperts} scales up the parametric memory compared to MoE, increases the granularity of sparsity, and avoids the cross-device communication overhead, but at the cost of high memory lookup cost during inference, and under-training issues of the large value memory. These challenges require sophisticated architectural modifications \citep{huang2025ultrasparsememorynetwork} and advanced system-level solutions \citep{berges2024memorylayersscale} to be overcome. 

Recently, Gemma-3n \citep{gemma3n2024} proposed Per Layer Embeddings (PLE) for small on-device models to \emph{complement} their limited parametric capacity with token-indexed sparse parametric memory. However, they do not dispose of original FFN modules, and use a much lower-dimensional PLE only to modulate the FFN output in each layer. These embedding tables are accommodated in fast storage, outside GPU HBM memory to accommodate larger batch sizes and enable fast prefetching.

%% file: sections/conclusion.tex
\section{Conclusion}
This work introduced \textsc{STEM}, a static, token-indexed design that replaces the FFN up-projection with a layer-local embedding lookup.This decouples parametric capacity from per-token compute and cross-device communication, yielding lower per-token FLOPs and fewer parameter accesses, and enabling CPU offload with asynchronous prefetch. Empirically, STEM trains stably despite extreme sparsity (compared to fine-grained MoE variants), improves accuracy over dense baselines, and exhibits higher effective memory capacity via a large-angular-spread embedding space. More importantly, it achieves all these benefits while being more interpretable. It also strengthens long-context performance by activating more distinct parameters as sequence length grows, providing practical test-time capacity scaling.
 

%% file: sections/appendix.tex
\section{Appendix}

\subsection{Additional Benchmarks}
Interestingly, our additional experiments on more contextual reasoning-heavy tasks show that STEM's contextual reasoning skill is better than that of the dense baseline. To directly probe reasoning beyond parametric knowledge, we evaluate 1B-scale baseline and STEM models on \emph{BIG-Bench Hard} (BBH) \citep{suzgun2022challengingbigbenchtaskschainofthought}, \emph{MuSR}\citep{sprague2024musrtestinglimitschainofthought}, and the \emph{LongBench}\citep{bai2024longbenchbilingualmultitaskbenchmark} multi-hop reasoning and code-understanding subsets. BBH is a collection of diverse, challenging tasks designed to require multi-step and compositional reasoning. MuSR requires the model to track entities and constraints over a long narrative before answering a question. The LongBench multi-hop subset tests reasoning across multiple passages, while the code-understanding subset evaluates comprehension of complex code snippets. As shown in Table~\ref{tab:context-reasoning}, STEM consistently outperforms the dense baseline on BBH, MuSR, and on LongBench multi-hop and code-understanding tasks across all context-length ranges, indicating that STEM does not impair contextual reasoning and can in fact improve it.

\begin{table}[h]
\centering
\small
\caption{\textbf{Contextual reasoning ability.} Contextual reasoning benchmarks for 1B-scale models. LongBench scores are averaged over tasks within each context-length range.}
\label{tab:context-reasoning}
\setlength{\tabcolsep}{4pt}
\begin{tabular}{lccccccccc}
\toprule
Model & BBH & MuSR 
& \multicolumn{3}{c}{LongBench Multi-hop} 
& \multicolumn{3}{c}{LongBench Code} \\
& & 
& $<4$k & 4--8k & $\ge 8$k 
& $<4$k & 4--8k & $\ge 8$k \\
\midrule
Baseline & 24.87 & 35.85 
& 5.72 & 6.20 & 6.19 
& 45.37 & 44.64 & 41.30 \\
STEM     & 27.55 & 36.38 
& 10.20 & 8.63 & 7.82 
& 52.68 & 52.53 & 49.60 \\
\bottomrule
\end{tabular}
\end{table}

\subsection{Additional Long-context Evaluation}
\label{App:longbench}
Apart from the synthetic task Needle-in-a-haystack, we further evaluate STEM on LongBench, a long-context benchmark that spans six task categories, including single- and multi-document question answering, summarization, few-shot learning, synthetic tasks, and code completion. We group test examples by context length and report the average scores in each regime. As shown in Table~\ref{tab:longbench-length-avg}, the 1B STEM model consistently matches or outperforms the 1B dense baseline across all context-length ranges, indicating that its long-context capabilities extend beyond synthetic tasks.

\begin{table}[h]
\centering
\small
\caption{\textbf{Long-context ability.} LongBench results (average across tasks) for 1B models, grouped by context length.}
\label{tab:longbench-length-avg}
\setlength{\tabcolsep}{6pt}
\begin{tabular}{lccccccc}
\toprule
Model & 0--2k & 2--4k & 4--6k & 6--8k & 8--10k & 10--12k & 12k+ \\
\midrule
Base & 24.0 & 23.8 & 22.1 & 22.3 & 21.9 & 21.1 & 23.5 \\
STEM & 27.6 & 27.6 & 24.4 & 22.7 & 23.0 & 21.7 & 24.2 \\
\bottomrule
\end{tabular}
\end{table}

\subsection{Additional Architecture Ablation Study}
\label{sec: additional arch}
To study the optimality of our STEM design principles, we further experiment with additional architecture alternatives.
\subsubsection{STEM$^\dagger$}
\label{sec:stem++}
STEM uses strictly fewer active parameters, and FLOPs for each token. And because of the architectural bias, STEM is susceptible to some loss of contextual learning ability. We also introduce a hybrid variant of STEM, which retains the up projection matrix in FFN, but complements with an additive token-specific modulation. Concretely, the new variant STEM$^\dagger$ computes the FFN output as follows,
\vspace{-0.25em}
\begin{equation}
  \mathbf{y}_{\ell}
  \;=\; \mathbf{W}_{\ell}^{d}\!\left(
      \mathrm{SiLU}\!\big(\mathbf{W}_{\ell}^{g} \mathbf{x}_{\ell}\big)
      \;\odot\;
      \left(\mathbf{W}_{\ell}^{u} \mathbf{x}_{\ell} + \mathbf{U}_{\ell}[t]\right)
  \right),
\end{equation}